\documentclass{article}
\usepackage{microtype}
\usepackage{graphicx}
\usepackage{subcaption}
\usepackage{booktabs} 
\usepackage[usenames,dvipsnames]{xcolor}
\usepackage{amsmath,amsfonts,amssymb,bm}
\usepackage{graphics,float}
\newcommand{\scalar}[2]{\langle #1 , #2 \rangle}
\def\eqref#1{equation~\ref{#1}}
\def\1{\bm{1}}

\def\va{{\bm{a}}}

\def\vg{{\bm{g}}}
\def\vh{{\bm{h}}}

\def\vu{{\bm{u}}}
\def\vv{{\bm{v}}}
\def\vw{{\bm{w}}}
\def\vx{{\bm{x}}}
\def\vy{{\bm{y}}}
\def\vz{{\bm{z}}}

\def\mA{{\bm{A}}}
\def\mB{{\bm{B}}}
\def\mC{{\bm{C}}}
\def\mD{{\bm{D}}}

\def\mL{{\bm{L}}}
\def\mM{{\bm{M}}}

\def\mT{{\bm{T}}}

\DeclareMathAlphabet{\mathsfit}{\encodingdefault}{\sfdefault}{m}{sl}
\SetMathAlphabet{\mathsfit}{bold}{\encodingdefault}{\sfdefault}{bx}{n}

\newcommand{\R}{\mathbb{R}}

\DeclareMathOperator*{\argmin}{arg\,min}

\def\alphab{\boldsymbol\alpha}
\def\betab{\boldsymbol\beta}

\def\G{\pi}
\def\GG{\boldsymbol\G}

\def\a{{\bf a}}

\def\h{{\bf h}}

\def\Mbf{{\bf M}}

\def\L{{\bf L}}
\def\R{{\mathbb{R}}}
\def\Abf{{\mathbf{A}}}

\def\diag{{\text{diag}}}

\def\vec{{\text{vec}}}
\def\tr{{\text{tr}}}
\def\one{{\mathbf{1}}}

\newcommand{\xbf}{\mathbf{x}}

\newcommand{\ybf}{\mathbf{y}}

\newcommand{\qbf}{\mathbf{q}}

\newcommand{\ie}{\textit{i.e.}}

\newcommand{\C}{\mathbf{C}}

\newcommand{\gw}{GW}
\newcommand{\fgw}{FGW}

\newtheorem{proposition}{Proposition}
\newtheorem{lemma}{Lemma}
\newtheorem{theorem}{Theorem}

\usepackage{comment}

\usepackage{hyperref}
\usepackage[accepted]{icml2021}
\usepackage{graphicx}

\makeatletter
\newcommand*\bigcdot{\mathpalette\bigcdot@{.5}}
\newcommand*\bigcdot@[2]{\mathbin{\vcenter{\hbox{\scalebox{#2}{$\m@th#1\bullet$}}}}}
\makeatother

\begin{document}
	
	\twocolumn[
	\icmltitle{Online Graph Dictionary Learning}
	\begin{icmlauthorlist}
		\icmlauthor{C\'{e}dric Vincent-Cuaz}{maasai}
		\icmlauthor{Titouan Vayer}{ens}
		\icmlauthor{R\'{e}mi Flamary}{pol}
		\icmlauthor{Marco Corneli}{maasai,smi}
		\icmlauthor{Nicolas Courty}{ubs}
	\end{icmlauthorlist}
	\icmlaffiliation{ens}{ENS\,de\,Lyon,\,LIP\,UMR\,5668,\,Lyon,\,France}
	\icmlaffiliation{pol}{Ecole\,Polytechnique,\,CMAP,\,UMR\,7641,\,Palaiseau,\,France}
	\icmlaffiliation{ubs}{Univ.Bretagne-Sud,\,CNRS,\,IRISA,\,Vannes,\,France}
	\icmlaffiliation{maasai}{Univ.C{\^o}te\,d{'}Azur,\,Inria,\,CNRS,\,LJAD,\,Maasai,\,Nice,\,France}
	\icmlaffiliation{smi}{Univ.C{\^o}te\,d{'}Azur,\,Center\,of\,Modeling,\,Simulation\,\&\,Interaction,\\Nice,\,France\\}
	\icmlcorrespondingauthor{C\'{e}dric Vincent-Cuaz}{cedric.vincent-cuaz@inria.fr}
	\icmlkeywords{Machine Learning, ICML}
	\vskip 0.3in
	]
	\printAffiliationsAndNotice{}
	
	\begin{abstract}
		Dictionary learning is a key tool for representation learning, that explains the
		data as linear combination of few basic elements. Yet, this analysis is not
		amenable in the context of graph learning, as graphs usually belong to different
		metric spaces. We fill this gap by proposing a new online Graph Dictionary
		Learning approach, which uses the Gromov Wasserstein divergence for the data
		fitting term. In our work, graphs are encoded through their nodes' pairwise relations 
		and modeled as convex combination of graph
		atoms, {\em i.e.} dictionary elements, estimated thanks to  an online stochastic
		algorithm, which operates on a dataset of unregistered graphs with potentially
		different number of nodes. Our approach naturally extends to labeled graphs, and
		is completed by a novel upper bound  that can be used as a fast approximation of
		Gromov Wasserstein in the embedding space. We provide numerical evidences
		showing the interest of our approach for unsupervised embedding of graph
		datasets and for online graph subspace estimation and tracking.
		
	\end{abstract}
	\section{Introduction}
	
	{The question of how to build machine learning algorithms able to go beyond
		vectorial data and to learn from structured data such as graphs has been of great
		interest in the last decades. Notable applications can be found in molecule
		compounds} \citep{kriege-recognizing-2018}, brain connectivity
	\citep{ktena-distance-2017}, social networks \citep{yanardag-deep-2015}, time
	series \citep{cuturi-soft-dtw-2018}, trees \citep{day-optimal-1985} {or} images
	\citep{harchaoui-image-2007,bronstein2017geometric}. %, etc.
	{Designing good representations for these data is challenging, as their
		nature is by essence non-vectorial, and requires dedicated modelling of their representing structures. Given
		sufficient data and labels, end-to-end approaches with neural networks have
		shown great promises in the last years \cite{wu2020comprehensive}. In this work, we focus on the unsupervised representation learning problem, where
		the entirety of the data might not be known beforehand, and is rather produced
		continuously by different sensors, and available through streams. {In this setting, tackling the non-stationarity of the underlying generating process is challenging~\citep{Ditzler2015}.  Good examples can be found, for instance, in the context of 
			dynamic functional connectivity~\citep{Heitmann2018} or network
			science~\citep{masuda2020guide}. As opposed to recent approaches focusing on dynamically varying graphs in online or continuous learning ~\citep{yang2018bandit,vlaski2018online,Wang2020StreamingGN},
we rather suppose in this work that \emph{distinct} graphs are made progressively available \cite{Zambon,Grattarola}. This setting is particularly challenging as the structure, the attributes or the number of nodes of each graph observed at a time step can differ from the previous ones. We propose to tackle this problem by learning a linear representation of graphs with online dictionary learning.}
		\vspace{1.5mm}
		\paragraph{Dictionary Learning (DL)}
		
		Dictionary Learning \citep{mairal2009online,schmitz2018wasserstein} is a field of unsupervised learning that aims at estimating a linear representation of the data, \ie\ to learn a linear subspace defined by the span of a family of vectors, called \emph{atoms}, which constitute a \emph{dictionary}. These atoms are inferred from the input data by minimizing a reconstruction error. These representations have been notably used in statistical frameworks such as data clustering \citep{ng-spectral-2002},
		recommendation systems \citep{bobadilla2013recommender} or dimensionality
		reduction  \citep{candes-robust-2009}. While DL methods mainly
		focus on vectorial data, 
		
		it is of prime interest to investigate flexible and interpretable
		factorization models applicable to \emph{structured data}. We also consider the dynamic or time varying version of the problem, where the data generating process may exhibit non-stationarity over time, yielding a problem of subspace change or tracking (see {\em e.g.} ~\cite{narayanamurthy2018nearly}), where one wants to monitor changes in the subspace best describing the data. In this work, we rely on optimal transport as a fidelity term to compare these structured data. 
		
		\paragraph{Optimal Transport for structured data}
		Optimal Transport (OT) theory provides a set of methods for comparing probability distributions, using, \emph{e.g.} the well-known Wasserstein
		distance \citep{villani-topics-2003}. 
		It has been notably} used by the machine learning community {in the
		context of distributional unsupervised learning}
	~\citep{arjovsky2017wasserstein, schmitz2018wasserstein,peyre-computational-2020}. Broadly speaking the interest of OT lies in its
	ability to provide correspondences, or relations, between sets of points.
	Consequently, it has recently garnered attention for learning tasks where
	the points are described by graphs/structured data (see  \emph{e.g.}
	\cite{DBLP:conf/aaai/NikolentzosMV17,maretic2019got,Togninalli19,xu2019scalable,vayer-optimal-nodate,barbe:hal-02795056}).
	One of the key ingredient in this case is to rely on the so called
	Gromov-Wasserstein (GW) distance
	\citep{memoli-gromovwasserstein-2011,sturm2012space} which is an OT
	problem adapted to the scenario in which the supports of the probability
	distributions lie in different metric spaces. The GW distance is
	particularly suited for comparing \emph{relational data}
	\cite{peyre2016gromov,solomon_entropic_2016} and, in a graph context, is
	able to find the relations between the nodes of two graphs when their
	respective structure is encoded through the pairwise relationship between
	the nodes in the graph. GW has been further studied for weighted directed
	graphs in \citep{chowdhury2019gromov} and has been extended to labeled
	graphs thanks to the Fused Gromov-Wasserstein (FGW) distance in
	\citep{vayer-fused-2018}. Note that OT divergences as losses for linear and
	non-linear DL over vectorial data have already been proposed in
	\citep{BTSSPP15,rolet2016fast, schmitz2018wasserstein} but the case of
	structured data remains quite unaddressed. A non-linear DL approach for
	graphs based on GW was proposed in \citep{xu_gromov-wasserstein_2019} but
	suffers from a lack of interpretability and high computational complexity
	(see discussions in Section \ref{sec:related_work}). To the best of our knowledge, a linear
	counterpart does not exist for now.

	\begin{figure}[!t]
		\includegraphics[width=\columnwidth]{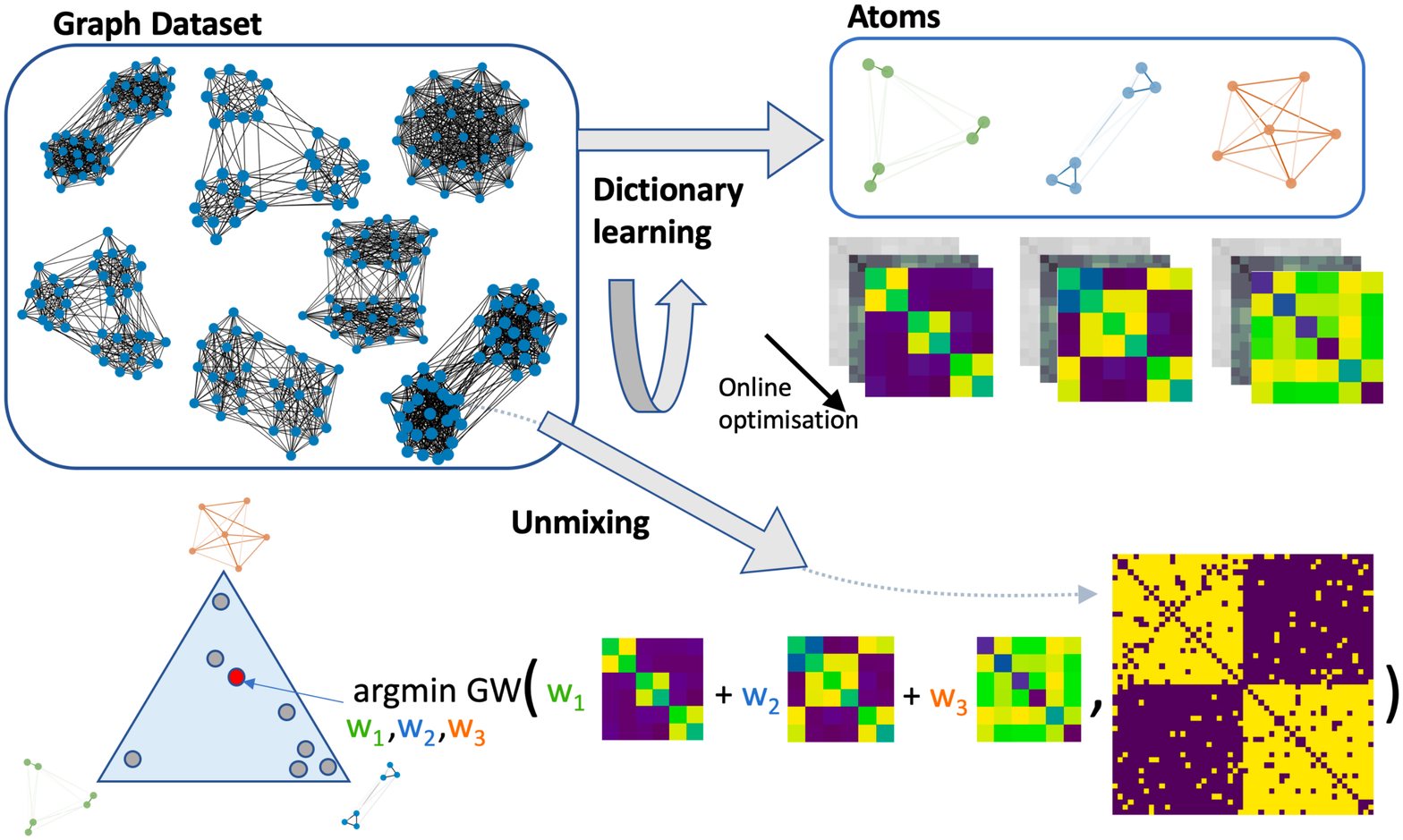}\vspace{-8mm}
		\caption{ From a dataset of graphs with different number of nodes, our method builds
			a dictionary of graph atoms with an online procedure. It uses the
			Gromov-Wasserstein distance as data fitting term between a  convex
			combination of the atoms and a pairwise relations representation for graphs
			from the dataset.}
		\label{fig:f1}
		\vspace{-5mm}
	\end{figure}
	
	\paragraph{Contributions}
	In this paper we use OT distances between structured data to design a linear and online DL for undirected graphs. Our proposal is depicted in Figure~\ref{fig:f1}. It
	consists in a new factorization model for undirected graphs optionally having node attributes relying on (F)GW distance as data fitting term. We propose an online 
	stochastic algorithm to learn the dictionary which scales to large real-world
	data (Section \ref{subsec:algos}), and uses extensively novel derivations of
	sub-gradients of the (F)GW distance (Section \ref{subsec:GW_subgrad}).  An
	unmixing procedure projects the graph in an embedding space defined {\em w.r.t.}
	the dictionary (Section \ref{subsec:GU}). Interestingly enough, we prove that
	the GW distance in this embedding is upper-bounded by a Mahalanobis distance
	over the space of \emph{unmixing} weights, providing a reliable and fast
	approximation of GW (Section \ref{subsec:GW_all}). Moreover, this approximation
	defines a proper kernel that can be efficiently used for clustering and
	classification of graphs datasets (sections
	\ref{subsec:simu}-\ref{subsec:real1}). We empirically demonstrate the relevance
	of our approach for online subspace estimation and subspace tracking by
	designing streams of graphs over two datasets (Section \ref{subsec:real2}).
	
	\paragraph{Notations}
	The simplex of histograms with $N$ bins is $\Sigma_N := \left\{\mathbf{h}\in
	\mathbb{R}^+_N| \sum_i h_i = 1 \right\}$. Let
	denote  $S_N(\R)$ the set of symmetric matrices in $\R^{N\times N}$. The Euclidean norm is denoted as $\|.\|_2$ and $\scalar{.}{.}_F$ the Frobenius inner product. We denote the gradient of a function $f$ over $\vx$ at $\vy$ in a stochastic context by $\widetilde{\nabla}_{\vx}f(\vy)$. The number nodes in a graph is called  the \emph{order} of the graph.

	\section{Online Graph Dictionary Learning }
	
	\subsection{(Fused) Gromov-Wasserstein for graph similarity}
	\label{subsec:GW_all}
	A graph $G^X$ with $N^X$ nodes, can be regarded as a tuple $(\mC^{X},\h^{X})$ where $\mC^{X} \in \R^{N^{X} \times N^{X}}$ is a matrix that
	encodes a notion of similarity between nodes and $\h^{X} \in \Sigma_{N^{X}}$ is
	a histogram, or equivalently a vector of weights  which models the relative importance of the nodes within the graph. Without any prior knowledge uniform weights can be chosen so that $\h^{X}=\frac{1}{N^{X}}\mathbf{1}_{N^{X}}$. The matrix $\mC^{X}$ carries the neighbourhood information of the nodes and, depending on
	the context, it may designate the adjacency matrix, the Laplacian matrix \cite{maretic2019got} or the matrix of the shortest-path distances between the nodes~\citep{bavaud2010euclidean}. Let us now consider two graphs $G^X=(\mC^{X},\h^{X})$ and $G^Y=(\mC^{Y},\h^{Y})$, {of potentially different orders (\emph{i.e} $N^X \neq N^Y$)}. The $GW_2$ distance between $G^X$ and $G^Y$ is defined as the result of the following optimization problem:
	\begin{equation}
	\label{eq:gwdef}
	\min_{\mT \in \mathcal{U}(\vh^X,\vh^Y)} \sum_{ijkl} \left( C^X_{ij}-C^Y_{kl}\right)^2T_{ik}T_{jl} 
	\end{equation}
	where $\mathcal{U}(\vh^X,\vh^Y):= \{\mT \in \R_{+}^{N^X\times N^Y}|\mT\mathbf{1}_{N^{Y}}=\vh^X, \mT^T\mathbf{1}_{N^{X}}=\vh^Y\}$ is the set of couplings between $\vh^X,\vh^Y$. 
	The optimal coupling $\mT$ of the GW problem acts as a probabilistic matching of nodes which tends to associate pairs of nodes that have similar pairwise relations in $\mC^X$ and $\mC^{Y}$, respectively. In the following we denote by $GW_2(\mC^{X},\mC^{Y},\vh^X,\vh^Y)$ the optimal value of \eqref{eq:gwdef} or  by $GW_2(\mC^{X},\mC^{Y})$ when the weights are uniform.
	
	The previous framework can be extended to graphs with node attributes
	(typically $\R^{d}$ vectors). In this case we  use the Fused
	Gromov-Wasserstein distance (FGW) \cite{vayer-fused-2018,vayer-optimal-nodate}
	instead of GW. More precisely, a labeled graph $G^X$ with $N^X$ nodes can be
	described this time as a tuple $G^X=(\mC^X,\mA^X, \vh^X)$ where $\mA^X \in \R^{N^X \times
		d}$ is the matrix of all features. Given two labeled graphs $G^X$ and $G^Y$,
	FGW aims at finding an optimal coupling by minimizing an OT cost which is a
	trade-off of a Wasserstein cost between the features and a GW cost between the
	similarity matrices. For the sake of clarity, we detail our approach in the GW context and refer the reader to the supplementary material for its extension to FGW.
	
	\subsection{Linear embedding and GW unmixing}\label{subsec:GU}
	
	\paragraph{Linear modeling of graphs} We propose to model a graph as a weighted sum of pairwise relation
	matrices. More precisely, given a graph $G=(\mC,\vh)$ and a \emph{dictionary}
	$\{\overline{\mC}_{s}\}_{s \in [S]}$ we want to find a linear representation $\sum_{s
		\in [S]} w_{s} \overline{\mC}_{s}$ of the graph $G$, as faithful as
	possible. The dictionary is made of pairwise relation matrices of graphs with
	order $N$. Thus, each $\overline{\mC}_{s} \in S_N(\R)$  is called an
	\emph{atom}, and $\vw=(w_{s})_{s \in [S]} \in \Sigma_S$ is referred as
	\emph{embedding} and denotes the coordinate of the graph $G$ in the dictionary
	as illustrated in Fig.\ref{fig:f1}. We rely
	on the GW distance to assess the quality of our linear
	approximation and propose to minimize it to estimate its optimal embedding. In
	addition to being interpretable thanks to its linearity, we also propose to promote
	sparsity in the weights $\vw$ similarly to sparse coding \citep{chen2001atomic}.
	Finally note that, when the pairwise matrices $\mC$ are adjacency matrices
	and the dictionary atoms have components in $[0,1]$, the model
	$\sum_{s \in [S]} w_{s} \overline{\mC}_{s}$ provides a matrix whose components
	can be interpreted as probabilities of connection between the nodes.

	\paragraph{Gromov-Wasserstein unmixing}
	We first study the unmixing problem that consists in projecting a graph on the
	linear representation discussed above, \emph{i.e.} estimate the optimal
	embedding $\vw$ of a graph $G$.
	The unmixing problem can be expressed as the minimization of the GW
	distance between the similarity matrix associated to the graph and its linear representation in the dictionary:
	\begin{align}   
	\min_{\vw \in \Sigma_S}\quad GW^2_2\left(\mC, \sum_{s \in [S]} w_s \overline{\mC}_s\right) - \lambda \|\vw\|^2_2 \label{eq:unmix}
	\end{align}
	where $\lambda \in \mathbb{R}^{+}$ induces a \textbf{negative} quadratic regularization promoting sparsity on the
	simplex as discussed in \citet{li2016methods}. In order to solve the non-convex problem in \eqref{eq:unmix}, we propose to
	use a Block Coordinate Descent (BCD) algorithm \citep{tseng2001convergence}. 
	\begin{algorithm}[t]
		\caption{BCD for unmixing problem \ref{eq:unmix}}
		\label{alg:BCD1}
		\begin{algorithmic}[1]
			\STATE Initialize $\vw=\frac{1}{S}\mathbf{1}_S$
			\REPEAT
			\STATE Compute OT matrix $\mT$ of $GW_2^2(\mC, \sum_s w_s \overline{\mC}_s)$, with CG algorithm ~\citep[Alg.1 \& 2]{vayer-fused-2018}.
			\STATE Compute the optimal $\vw$ solving \eqref{eq:unmix} for a fixed
			$\mT$ with CG algorithm. 
			\UNTIL{convergence}
		\end{algorithmic}
	\end{algorithm}
	
	The BCD (Alg.\ref{alg:BCD1}) works by alternatively updating the OT matrix of the GW distance and the embeddings $\vw$. When $\vw$ is fixed the problem
	is a classical GW  {which is a non-convex quadratic program. We 
		solve it} using a Conditional Gradient (CG) algorithm
	\citep{jaggi2013revisiting} based on \citep{vayer-optimal-nodate}.  Note that the use of the exact GW instead of a regularized
	proxy allowed us to keep a sparse OT matrix as well as
	to  preserve ``high frequency'' components of the graph, as opposed to
	regularized versions of GW \citep{peyre2016gromov,solomon_entropic_2016,xu2019gromov} that promotes dense OT matrices
	and leads to smoothed/averaged pairwise matrices. For a fixed OT matrix $\mT$, the problem of finding $\vw$ is a non-convex
	quadratic program and can also be tackled with a CG algorithm.  Note that for non-convex
	problems the CG algorithm is
	known to converge to a local
	stationary point \citep{lacoste-julien-convergence-2016}. In practice, we
	observed a typical convergence of the CGs in a few tens of iterations. The BCD
	itself converges in less than 10 iterations.
	\paragraph{Fast upper bound for GW} Interestingly, when two graphs belong to
	the linear subspace defined by our dictionary, there exists a proxy of the GW distance using a
	dedicated Mahalanobis distance as described in the next propositon: 
	\begin{proposition}
		\label{prop:embed_graph}
		For two embedded graphs with embeddings $\vw^{(1)}$ and $\vw^{(2)}$, assuming they share the same weights $\vh$, the following inequality holds
		\begin{align}%\small
		GW_2\left(\sum_{s \in [S]} w^{(1)}_s \overline{\mC}_s, \sum_{s \in [S]} w^{(2)}_s \overline{\mC}_s\right) \leq \|\vw^{(1)} - \vw^{(2)}\|_\mM \label{eq:mah_gw}
		\end{align}
		where $M_{pq} = \scalar{\mD_{\vh}\overline{\mC}_p}{ \overline{\mC}_q\mD_{\vh}}_F$ and $\mD_\vh= diag(\vh)$. $\mM$ is a positive semi-definite matrix hence engenders a Mahalanobis distance between embeddings.
	\end{proposition}
	As detailed in the supplementary material, this upper bound is obtained by considering the GW cost between the linear models calculated using the admissible coupling $\mD_{\vh}$. The latter coupling assumes that both graph representations are aligned and therefore is a priori suboptimal. As such, this bound is not tight in general. However, when the embeddings are close, the optimal coupling matrix should be close to $\mD_{\vh}$ so that Proposition \ref{prop:embed_graph} provides a reasonable proxy to the GW distance into our embedding space. In practice, this upper bound can be used to compute efficiently pairwise kernel matrices or to do retrieval of closest samples (see numerical experiments).
	
	\subsection{Dictionary learning and online algorithm}\label{subsec:algos}
	Assume now that the dictionary $\{\overline{\mC}_s\}_{s \in [S]}$ is not known
	and has to be estimated from the data.
	We define a dataset of $K$ graphs $\left\{ G^{(k)} :
	(\mC^{(k)},\vh^{(k)}) \right\}_{k \in [K]}$. Recall that each graph $G^{(k)}$ of
	order $N^{(k)}$ is summarized by its pairwise relation matrix $\mC^{(k)} \in
	S_{N^{(k)}}(\R)$ and weights $\vh^{(k)} \in \Sigma_{N^{(k)}}$ over nodes, as
	described in Section \ref{subsec:GW_all}.
	The DL problem, that aims at estimating the optimal dictionary
	for a given dataset can be expressed as:
	\begin{align}
	\min_{\begin{smallmatrix}\{\vw^{(k)}\}_{k\in [K]} \\
		\{\overline{\mC}_s\}_{s \in [S]} \end{smallmatrix}} &\sum_{k=1}^K GW^2_2\left(\mC^{(k)},\sum_{s \in [S]} w^{(k)}_s \overline{\mC}_s\right)- \lambda \|\vw^{(k)}\|^2_2 \label{eq:dl}
	\end{align}
	where $\vw^{(k)} \in \Sigma_S, \overline{\mC}_s \in S_N(\R)$. Note that the optimization problem above is a classical sparsity promoting
	dictionary learning on a linear subspace but with the important novelty
	that the reconstruction error is computed with the GW distance. This allows us to learn a graphs
	subspace of fixed
	order $N$ using a dataset of graphs with various orders.
	The sum over the errors in \eqref{eq:dl} can be seen as an expectation and we
	propose to devise an online strategy to optimize the problem similarly to the
	online DL proposed in \citep{mairal2009online}. The main
	idea is to update the dictionary $\{\overline{\mC}_s\}_s$ with a stochastic
	estimation of the gradients on few dataset graphs (minibatch). At each
	stochastic update the unmixing problems are solved independently for each
	graph of the minibatch using a fixed dictionary 
	$\{\overline{\mC}_s\}_s$, {using the procedure described in Section
		\ref{subsec:GU}}. Then one can compute a gradient of the loss on the minibatch
	\emph{w.r.t} 
	$\{\overline{\mC}_s\}_s$ and proceed to a projected gradient step.  The
	stochastic update of the proposed
	algorithm is detailed in Alg.\ref{alg:GW1}. Note that it
	can be used on a finite dataset with possibly several epochs on the whole
	dataset or online in the presence of streaming graphs. We provide an example
	of such subspace tracking in Section \ref{subsec:real2}.  We will refer to our
	approach as GDL in the rest of the paper.

	\begin{algorithm}[t]
		\caption{GDL: stochastic update of atoms $\{\overline{\mC}_s\}_{s\in [S]}$}
		\label{alg:GW1}
		\begin{algorithmic}[1]
			\STATE Sample a minibatch of graphs $\mathcal{B} :=\{\mC^{(k)}\}_{k \in \mathcal{B}}$ .
			\STATE Compute optimal $\{(\vw^{(k)},\mT^{(k)})\}_{k \in [B]}$ by solving B independent unmixing problems with Alg.\ref{alg:BCD1}. 
			\STATE Projected gradient step with estimated gradients $\widetilde{\nabla}_{\overline{\mC}_s}$ (equation in supplementary), $\forall s \in [S]$: \vspace{-2mm}
			\begin{equation}
			\overline{\mC}_s \leftarrow Proj_{S_N(\R)}( \overline{\mC}_s - \eta_C \widetilde{\nabla}_{\overline{\mC}_s} )
			\end{equation}
		\end{algorithmic}
	\end{algorithm}
	
	\paragraph{Numerical complexity} The numerical complexity of GDL depends on the complexity of each update. The main computational bottleneck is the unmixing procedure that relies on multiple resolution of GW
	problems. The complexity of solving a GW with the {CG} algorithm between two graphs of order $N$ and $M$ and computing
	its gradient is dominated by  $\mathcal{O}\left(N^2
	M+M^2N\right)$ operations
	\cite{peyre2016gromov,vayer-fused-2018}.
	Thus given dictionary
	atoms of order $N$, the worst case complexity 
	can be only \textbf{quadratic} in
	the highest graph order in the dataset.  For instance, estimating embedding on dataset IMDB-M (see Section \ref{subsec:real1}) over 12 atoms takes on average $44$ ms per graph (on processor i9-9900K CPU 3.60GHz). {We refer the reader to the
		supplementary for more details.}
	Note that in addition to scale well to large datasets thanks to the stochastic
	optimization, our method also leads to important speedups when using the
	representations as input feature for other ML tasks. For instance, we can use the upper bound in
	\eqref{eq:mah_gw} to compute efficiently kernels between graphs instead of
	computing all pairwise GW distances.
	\paragraph{GDL on labeled graphs}{We can also define the same DL procedure for labeled graphs using the FGW distance. The unmixing part defined in \eqref{eq:unmix} can be adapted by considering a linear embedding of the similarity matrix \emph{and} of the feature matrix parametrized by the \emph{same} $\vw$. From an optimization perspective, finding the optimal coupling of FGW can be achieved  using a CG procedure so that Alg.\ref{alg:GW1} extends naturally to the FGW case. Note also that the upper bound
		of Proposition \ref{prop:embed_graph} can be generalized to this setting. This discussion is
		detailed in supplementary material.}
	
	\subsection{Learning the graph structure and distribution}
	\label{subsec:GW_subgrad}
	
	Recent researches have studied the use of potentially more general distributions
	$\vh$
	on the nodes of graphs than the naive uniform ones commonly used. \citep{xu2019scalable} empirically explored the use of distributions induced by
	degrees, such as parameterized power
	laws, $h_i = \frac{p_i}{\sum_i p_i}$, where $p_i = (deg(x_i)+a)^b$ with $a \in
	\R_+$ and $b \in [0,1]$. They demonstrated the interest of this approach but also highlighted how hard it is to calibrate, which {advocates for learning these distributions.}
	With this motivation, we extend our GDL model defined in
	equation \ref{eq:dl} and propose to learn atoms of the form $\{\overline{\mC}_s,\overline{\vh}_s\}_{s\in [S]}$. In this setting we have
	two independent dictionaries modeling the relative importance of the nodes with
	$\overline{\vh}_s \in \Sigma_N$, and their pairwise relations through $\overline{\mC}_s$. This
	dictionary learning problem reads:
	\begin{align}
	\min_{\substack{\{(\vw^{(k)},\vv^{(k)})\}_{k \in [K]}\\
			\{(\overline{\mC}_s,\overline{\vh}_s)\}_{s\in [S]}}} \sum_{k=1}^K & GW^2_2\left(\mC^{(k)},\widetilde{\mC}(\vw^{(k)}), \vh^{(k)}, \widetilde{\vh}(\vv^{(k)})\right) \nonumber\\
	& - \lambda \|\vw^{(k)}\|^2_2 - \mu \|\vv^{(k)}\|^2_2\label{eq:dl_h}
	\end{align}
	where $\vw^{(k)},\vv^{(k)} \in \Sigma_S$ are the structure and distribution embeddings and the linear models are defined as:
	\begin{equation*}
	\forall k,\
	\widetilde{\vh}(\vv^{(k)}) = \sum_s v^{(k)}_s\overline{\vh}_s,\quad  
	\widetilde{\mC}(\vw^{(k)}) = \sum_s w^{(k)}_s \overline{\mC}_s 
	\end{equation*}
	Here we exploit fully the GW
	formalism by estimating simultaneously the graph distribution $\widetilde{\vh}$
	and its geometric structure $\widetilde{\mC}$. Optimization problem \ref{eq:dl_h} can
	be solved by an adaptation of stochastic Algorithm \ref{alg:GW1}. We estimate
	the structure/node weights unmixings
	$(\vw^{(k)},\vv^{(k)})$ over a minibatch of graphs with a BCD (see Section
	\ref{subsec:algos}). Then we perform simultaneously a projected gradient step
	update of $\{\overline{\mC}_s\}_s$ and $\{\overline{\vh}_s\}_s$. More details are given in the supplementary.

	The optimization procedure above requires to have access to a gradient for
	the GW distance \emph{w.r.t.} the weights. To the best of our knowledge no
	theoretical results exists in the literature for finding such gradients. We
	provide below a simple way to compute a subgradient for GW weights from subgradients of the well-known Wasserstein distance:
	\begin{proposition}\label{eq:prop3}
		{Let $(\mC^1,\vh^1)$ and $(\mC^2,\vh^2)$ be two graphs. Let $\mT^{*}$ be an optimal coupling of the GW problem between $(\mC^1,\vh^1),(\mC^2,\vh^2)$. We define the following cost matrix $\mM(\mT^{*}) := \left(\sum_{kl}(C^1_{ik}- C^2_{jl})^2T^{*}_{kl} \right)_{ij}$. Let $\alphab^{*}(\mT^{*}),\betab^{*}(\mT^{*})$ be the dual variables of the following linear OT problem:}
		\begin{equation}
		\min_{\mT \in \mathcal{U}(\vh^1,\vh^2)}\langle \mM(\mT^{*}) ,\mT\rangle_{F}
		\end{equation}
		Then $\alphab^{*}(\mT^{*})$ (\textit{resp} $\betab^{*}(\mT^{*})$) is a subgradient of the function $GW_2^2(\mC^1,\mC^2,\ \bigcdot\ ,\vh^2)$ (\textit{resp} $GW_2^2(\mC^1,\mC^2,\vh^1, \ \bigcdot \ )$).
	\end{proposition}
	The proposition above shows that the subgradient of GW \textit{w.r.t.} the
	weights can be found by solving a linear OT problem which corresponds to a Wasserstein distance. The ground cost $\mM(\mT^{*})$ of this Wasserstein is moreover the gradient (\textit{w.r.t.} the couplings) of the optimal GW loss.
	Note that in practice the GW problem is solved with a CG algorithm which already
	requires to solve this linear OT problem at each iteration. In this way, after
	convergence, the gradient \textit{w.r.t.} the weights can be extracted for free
	from the last iteration of the CG algorithm. 
	The proof of proposition \ref{eq:prop3} is given in the supplementary material.
	
	\section{Related work and discussion \label{sec:related_work}}
	
	In this section we discuss the relation of our GDL framework with existing
	approaches designed to handle graph data. We first focus on existing contributions
	for graph representation in machine learning applications. Then, we discuss in more details the existing non-linear graph dictionary learning approach of \citep{xu_gromov-wasserstein_2019}.

	\paragraph{Graph representation learning} Processing of graph data
	in machine learning applications have traditionally been handled using implicit
	representations such as with graph kernels \citep{shervashidze09,vishwanathan2010graph}. Recent results have shown the
	interest of using OT based distances to measure graph similarities and to design new
	kernels~\citep{vayer-optimal-nodate,maretic2019got,chowdhury-generalized-2020}. However, one limit of kernel methods is
	that the representation of the graph is fixed \emph{a priori} and cannot be
	adapted to specific datasets. On the other hand, Geometric deep learning approaches \citep{bronstein2017geometric} attempt to learn
	the representation for structured data by means of deep learning~\citep{scarselli2008graph, perozzi2014deepwalk, niepert-learning-2016}. Graph Neural Networks \cite{wu2020comprehensive} have
	shown impressive performance for end-to-end supervised learning problems. Note
	that both kernel methods and many deep learning based representations for graphs suffer from the fundamental \emph{pre-image} problem, that prevents recovering actual
	graph objects from the embeddings. 
	Our proposed GDL aims at overcoming such a limit relying on an unmixing
	procedure that not only provides a simple vectorial representation on the
	dictionary but also allows a direct reconstruction of interpretable graphs (as
	illustrated in the experiments). A recent contribution potentially overcoming
	the pre-image problem  is~\citet{Grattarola}. In that paper, a variational
	autoencoder is indeed trained to embed the observed graphs into a constant
	curvature Riemannian manifold. The aim of that paper is to represent the graph
	data into a space where the statistical tests for change detection are easier.
	We look instead for a latent representation of the graphs that remains as
	interpretable as possible. As a side note, we point out that our GDL embeddings
	might be used as input for the statistical tests developed by
	\citep{Zambon,Zambon2019ChangePointMO} to detect stationarity changes in the
	stochastic process generating the observed graphs~(see for instance
	Figure~\ref{fig:online}) .
	\paragraph{Non-linear GW dictionary learning of graphs} 
	In a recent work,
	\citep{xu_gromov-wasserstein_2019} proposed a non-linear factorization of
	graphs using a regularized version of GW barycenters~\citep{peyre2016gromov} and
	denoted it as Gromov-Wasserstein Factorization (GWF). Authors propose to learn a dictionary $\{\overline{\mC_s}\}_{s\in[S]}$ by 
	minimizing over $\{\overline{\mC}_s\}_{s\in[S]}$ and $\{\vw^{(k)}\}_{k\in[K]}$ the
	quantity $\sum_{k=1}^{K} GW_2^{2}(\widetilde{\mB}(\vw^{(k)};
	\{\overline{\mC}_s\}_s),\mC^{(k)})$ where $\widetilde{\mB}(\vw^{(k)};\{\overline{\mC}_s\}_s ) \in \arg\min_{\mB} \sum_{s}
	w^{(k)}_s GW_2^{2}(\mB,\overline{\mC}_s)$ is a GW barycenter. The main difference between GDL and this
	work lies in the linear representation of the approximated graph that we adopt whereas
	\citep{xu_gromov-wasserstein_2019} relies on the highly non-linear Gromov
	barycenter. As a consequence, the unmixing requires solving a complex
	bi-level optimization problem that is computationally expensive. Similarly,
	reconstructing a graph from this embedding requires again the resolution of a
	GW barycenter, whereas our linear reconstruction process is immediate. 
	In Section~\ref{sec:exp}, we show that our GDL
	representation technique compares favorably to GWF, both in terms of numerical complexity and
	performance.
	
	\section{Numerical experiments}\label{sec:exp}
	This section aims at illustrating the behavior of the approaches introduced so far for both clustering (Sections \ref{subsec:simu}-\ref{subsec:real1})  and online subspace tracking (Section \ref{subsec:real2}).

	\paragraph{Implementation details} The base OT solvers that are used in the
	algorithms rely on the POT toolbox \citep{flamary2017pot}.
	For our experiments, we considered the Adam algorithm \citep{kingma2014adam} as
	an adaptive strategy for the update of the atoms with a fixed dataset, but used
	SGD with constant step size for the online experiments in Section~\ref{subsec:real2}. The code is available at ~\href{https://github.com/cedricvincentcuaz/GDL}{https://github.com/cedricvincentcuaz/GDL}.
	
	\subsection{GDL on simulated datasets}\label{subsec:simu}
	The GDL approach discussed in this section refers to~\eqref{eq:dl}.
	First we illustrate it on datasets simulated according to the well understood Stochastic Block Model~\citep[SBM,][]{holland1983stochastic,wang1987stochastic} and show that we can recover embeddings and dictionary atoms corresponding to the generative structure.
	
\begin{figure}[t]
		\centering
		\includegraphics[width=\linewidth]{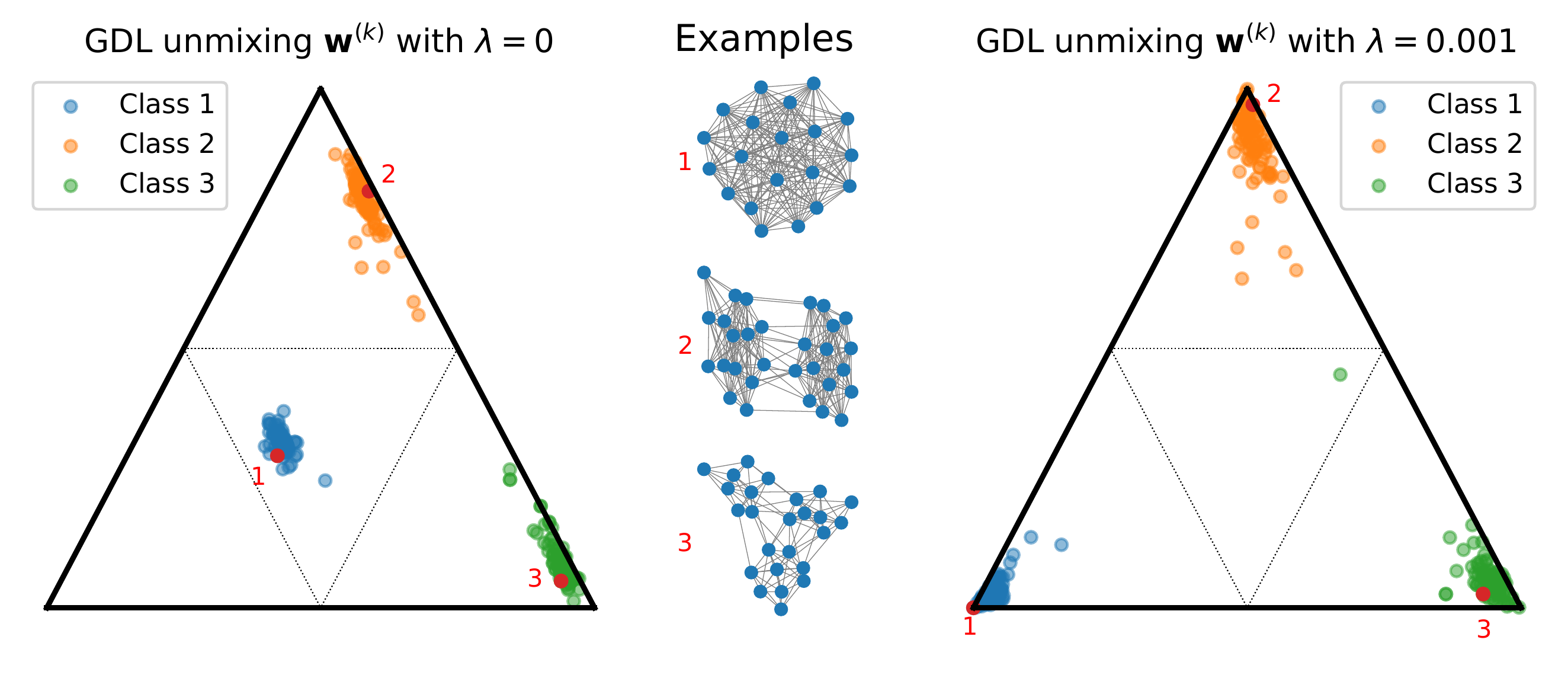}\vspace{-4mm}
		\caption{Visualizations of the embeddings of the graphs from $D_1$ with our GDL on 3 atoms. The positions on
			the simplex for the different classes are reported with no regularization
			(left) and sparsity promoting regularization (right). Three simulated graphs from $D_1$ are shown in the middle and their positions
			on the simplex reported in red.} \label{fig:simplex}
		\vspace{-5mm}
	\end{figure}

\begin{figure}[t]
		\centering
		\includegraphics[width=\linewidth]{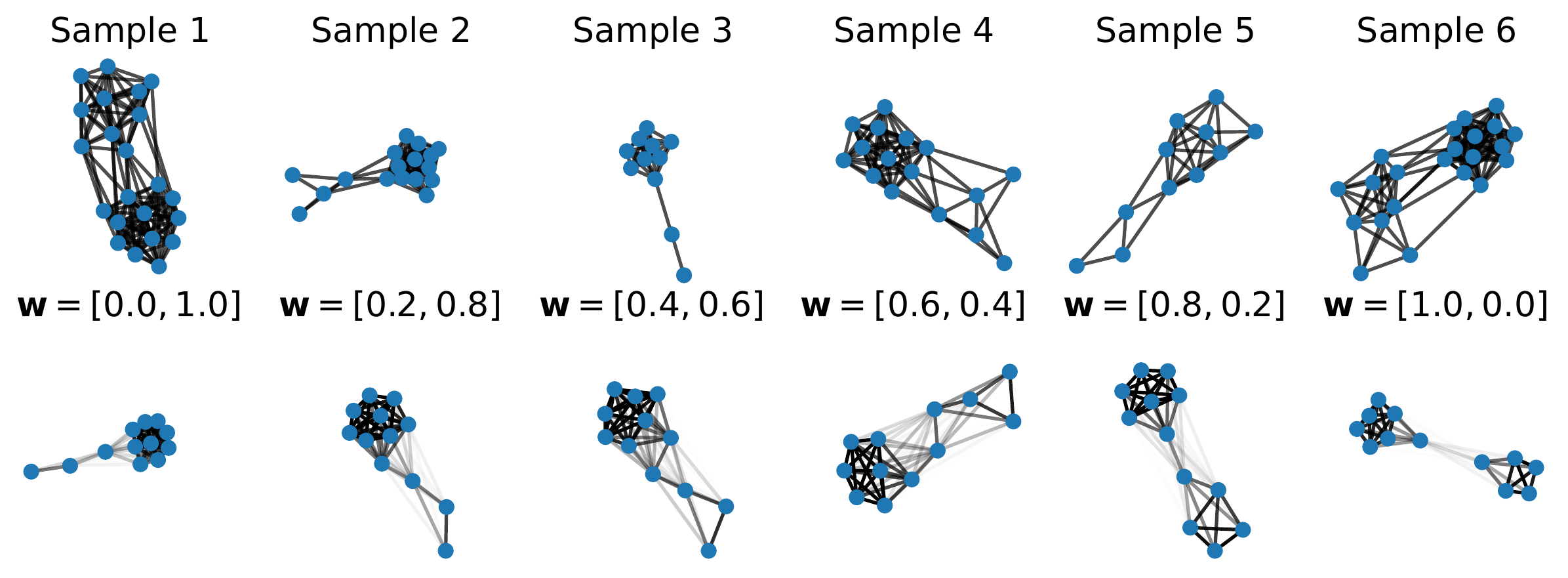}\vspace{-2mm}
		\caption{On the top, a random sample of real graphs from $D_2$ (two blocks). 
			On the bottom, reconstructed graphs as linear combination of two estimated atoms (varying proportions for each atom).} \label{fig:interp_toy}
	\end{figure}

	\paragraph{Datasets description}
	We consider two datasets of graphs, generated according to SBM, with various orders, randomly sampled in $\{10,15,...,60\}$ .
	The first scenario ($D_1$) adopts three different generative structures (also
	referred to as \emph{classes}): dense (no clusters), two clusters and three
	clusters~(see Figures~\ref{fig:simplex}). Nodes are assigned to
	clusters into equal proportions. For each generative structure 100 graphs are
	sampled.
	The second scenario $(D_2)$ considers the generative structure with two clusters,
	but with varying proportions of nodes for each block (see top of
	Figure~\ref{fig:interp_toy}), 150 graphs are simulated
	accordingly.
	In both scenarios we fix $p=0.1$ as the probability of inter-cluster connectivity and $1-p$ as the probability of intra-cluster connectivity. We consider adjacency matrices for representing the structures of the graphs in the datasets and uniform weights on the nodes.
	
	\paragraph{Results and interpretation} First we learn on dataset $D_1$ a dictionary
	of 3 atoms of order 6. The unmixing coefficients for the
	samples in $D_1$ are reported in Fig.~\ref{fig:simplex}. On the left, we see that the coefficients are not sparse on the simplex but the samples are clearly
	well clustered and graphs sharing the same class (i.e. color) are well separated.  
	When adding sparsity promoting regularization  (right part of the figure) the
	different classes are clustered on the corners of the simplex, thus suggesting
	that regularization leads to a more discriminant representation. The estimated atoms
	for the regularized GDL are reported on the top of Fig.~\ref{fig:f1} as both
	matrices $\overline{\mC}_s$ and their corresponding graphs.
	As it can be seen, the different SBM structures in $D_1$ are recovered.
	Next we estimate on $D_2$ a dictionary with 2 atoms of order 12. 
	The interpolation between the two estimated atoms for some samples is reported in Fig.~\ref{fig:interp_toy}.
	As it can be seen, $D_2$ can be modeled as a one dimensional manifold where the
	proportion of nodes in each block changes continuously. We stress that the grey links on the bottom of Figure~\ref{fig:interp_toy} correspond to the entries of the reconstructed adjacency matrices. Those entries are in $[0,1]$, thus encoding a probability of connection (see Section~\ref{subsec:GU}). 
	The darker the link, the higher the probability of interaction between the corresponding nodes. The possibility of generating random graphs using
	these probabilities opens the door to future researches.
	
\begin{figure}[t]
		\centering
		\includegraphics[width=\linewidth]{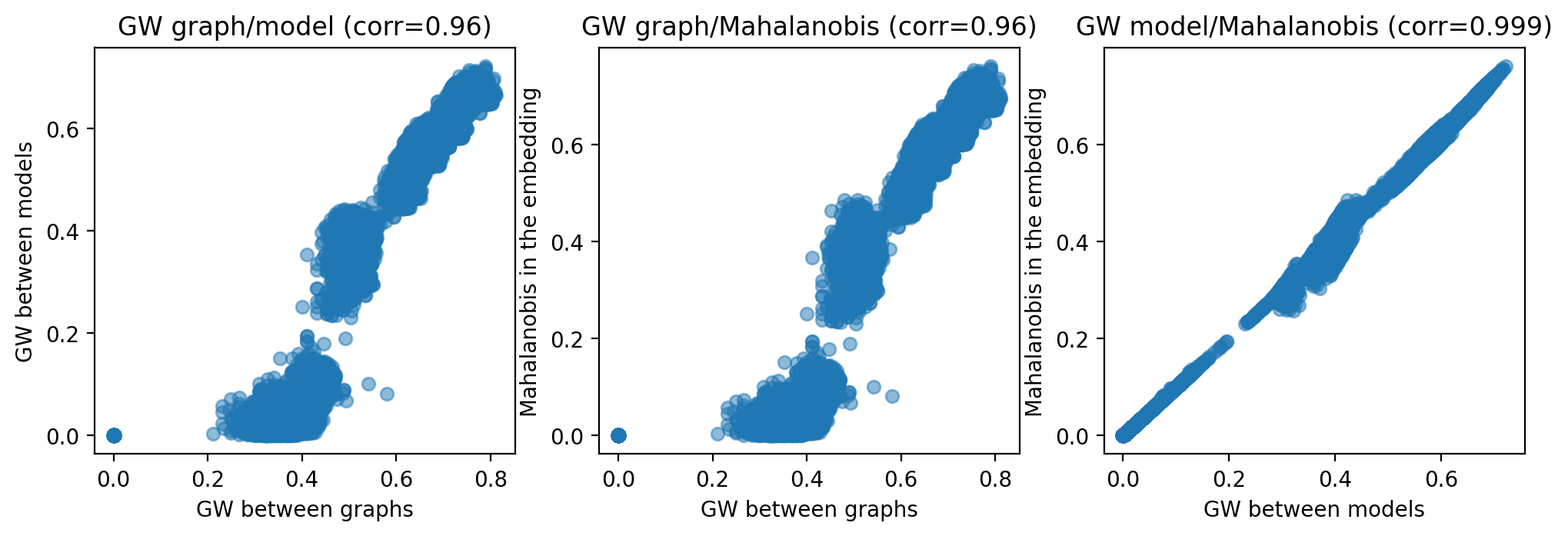}\vspace{-1mm}
		\caption{Plot of the pairwise distances in $D_1$ and their Pearson 
			correlation coefficients. GW distance between graphs versus its counterpart between the embedded graphs (left). GW distance between graphs versus  Mahalanobis distance between the embeddings (middle). GW distance between the embedded graphs versus Mahalanobis between the corresponding embeddings (right). } \label{fig:dist}
	\end{figure}

	We evaluate in Fig.~\ref{fig:dist} the
	quality of the Mahalanobis upper bound in~\eqref{eq:mah_gw} as a proxy for the GW distance on
	$D_1$. 
	On the left, one can see that the linear model allows us to recover the true GW distances between graphs most of the time.
	Exceptions occur for samples in the same class (i.e.  "near" to each other in terms of GW distance).
	The right part of the figure shows that the
	correlation between the Mahalanobis upper bound (cf. Proposition~\ref{prop:embed_graph}) and the GW distance between the
	embedded graphs is nearly perfect (0.999). This proves that our proposed upper bound
	provides a nice approximation of the GW distance between the input graphs, with a correlation of 0.96 (middle of the figure), at a much lower computational cost.  
	
	\subsection{GDL on real data for clustering and classification}\label{subsec:real1} 
	We now show how our \emph{unsupervised} GDL procedure can be used to find meaningful representations for well-known graph classification datasets. The knowledge of the classes will be
	employed as a ground truth to validate our estimated embeddings in \emph{clustering} tasks. 
	For the sake of completeness, in supplementary material we also report the supervised
	classification accuracies of some recent supervised graph \emph{classification}
	methods (e.g. GNN, kernel methods) showing that our DL and embedding is
	also competitive for classification.
	\paragraph{Datasets and methods} We considered well-known benchmark datasets divided into three categories: i) IMDB-B and IMDB-M~\citep{yanardag-deep-2015} gather graphs without node attributes derived from social networks; ii) graphs with discrete attributes representing chemical compounds from MUTAG~\citep{debnath1991structure} and cuneiform signs from PTC-MR~\citep{krichene2015efficient}; iii) graphs with real vectors as attributes, namely  BZR, COX2~\citep{sutherland2003spline}
	and PROTEINS, ENZYMES~\citep{borgwardt2005shortest}. We benchmarked our models for clustering tasks with the following state-of-the-art OT models: 
	i) GWF ~\citep{xu_gromov-wasserstein_2019}, using the proximal point algorithm  detailed in that paper and exploring two configurations, i.e. with either fixed atom order (GWF-f) or random atom order (GWF-r, default for the method); ii) GW k-means (GW-k)  which is a k-means using GW distances and GW barycenter \citep{peyre2016gromov};
	iii) Spectral Clustering (SC) of ~\citep{shi2000normalized,stella2003multiclass} applied to the pairwise GW distance matrices or the pairwise FGW distance matrices for graphs with attributes. We complete these clustering evaluations with an ablation study of the effect of the negative quadratic regularization proposed with our models. As introduced in \eqref{eq:dl}, this regularization is parameterized by $\lambda$, so in this specific context we will distinguish GDL ($\lambda=0$) from $\text{GDL}_{\lambda}$ ($\lambda>0$).
	
	\begin{table*}[t!]
		\caption{Clustering: Rand Index computed for benchmarked approaches on real datasets.}
		\label{tab:clustering}
		\begin{center}
			\scalebox{0.81}{
				
				\begin{tabular}{|c|c|c|c|c|c|c|c|c|}
					\hline
					& \multicolumn{2}{|c|}{NO ATTRIBUTE} & \multicolumn{2}{|c|}{DISCRETE ATTRIBUTES} &\multicolumn{4}{c|}{REAL ATTRIBUTES} \\
					\hline
					MODELS & IMDB-B & IMDB-M& MUTAG& PTC-MR & BZR & COX2 & ENZYMES & PROTEIN\\
					\hline
					GDL \text{ } (ours) & 51.32(0.30) & 55.08(0.28) & 70.02(0.29) & 51.53(0.36) & 62.59(1.68) & 58.39(0.52) & 66.97(0.93) & 60.22(0.30)\\
					$\text{GDL}_\lambda $ (ours) & $\mathbf{51.64(0.59)}$ & 55.41(0.20) & $\mathbf{70.89(0.11)}$ & $\mathbf{51.90(0.54)}$ & $\mathbf{66.42(1.96)}$ & $\mathbf{59.48(0.68)}$ & 66.79(1.12) & $\mathbf{60.49(0.71)}$\\ \hline 
					GWF-r &51.24 (0.02) & $\mathbf{55.54(0.03)}$ & 68.83(1.47) & 51.44(0.52) & 52.42(2.48) & 56.84(0.41)  & $\mathbf{72.13(0.19)}$& 59.96(0.09)\\
					GWF-f &50.47(0.34)& 54.01(0.37)& 58.96(1.91)& 50.87(0.79) & 51.65(2.96) & 52.86(0.53) & 71.64(0.31) & 58.89(0.39)\\ 
					GW-k &50.32(0.02) & 53.65(0.07)& 57.56(1.50) & 50.44(0.35) & 56.72(0.50) & 52.48(0.12) & 66.33(1.42) & 50.08(0.01) \\
					SC & 50.11(0.10) & 54.40(9.45) &50.82(2.71) & 50.45(0.31) & 42.73(7.06) & 41.32(6.07) & 70.74(10.60)& 49.92(1.23) \\ 
					\hline
			\end{tabular}}
		\end{center}
	\end{table*}
	
	\paragraph{Experimental settings} For the datasets with attributes involving FGW, we tested 15 values of the trade-off parameter $\alpha$ via a logspace search in $(0,0.5)$ and symmetrically
	$(0.5,1)$ and select the one minimizing our objectives.
	For our GDL methods as well as for GWF, a first step consists into learning the atoms.
	A variable number of $S=\beta k$ atoms is tested, where $k$ denotes the number of classes and $\beta \in \{2,4,6,8\}$, with a uniform number of atoms per class. 
	When the order $N$ of each atom is fixed, for GDL and GWF-f, it is set to the
	median order in the dataset. The atoms are initialized by randomly sampling graphs from
	the dataset with corresponding order. We tested 4 regularization coefficients for both methods.
	
	The embeddings $\vw$ are then computed and used as input for a
	k-means algorithm. However, whereas a standard Euclidean distance is used to
	implement k-means over the GWFs embeddings, we use the Mahalanobis
	distance from Proposition~\ref{prop:embed_graph} for the k-means clustering of the GDLs embeddings. Unlike
	GDL and GWF, GW-k and SC do not require any embedding learning step. Indeed,
	GW-k directly computes (a GW) k-means over the input graphs and SC is applied to
	the  GW distance matrix obtained from the input graphs. The cluster assignments
	are assessed by means of Rand Index ~\citep[RI,][]{rand1971objective}, computed
	between the true class assignment (known) and the one estimated by the different
	methods. For each parameter configuration (number of atoms, number of nodes and
	regularization parameter)
	we run each experiment five times, independently, with different random initializations. The mean RI was computed over the random
	initializations and the dictionary configuration leading to the highest RI was
	finally retained.
	
	\paragraph{Results and interpretation} Clustering results can be seen in Table
	\ref{tab:clustering}. The mean RI and its standard deviation are reported for each dataset and method.
	Our model outperforms or is at least comparable to the state-of-the-art OT based approaches for most of the datasets. Results show that the negative quadratic regularization proposed with our models brings additional gains in performance. Note that for this benchmark, we considered a fixed batch size for learning our models on labeled graphs, which turned out to be a limitation for the dataset ENZYMES. Indeed, comparable conclusions regarding our models performance have been observed by setting a higher batch size for this latter dataset and are reported in the supplementary material.
	This might be due to both a high number of heterogeneous classes and a high structural diversity of labeled graphs inside and among classes.
	
\begin{figure}[t]
		\centering
		\includegraphics[width=1.04\linewidth]{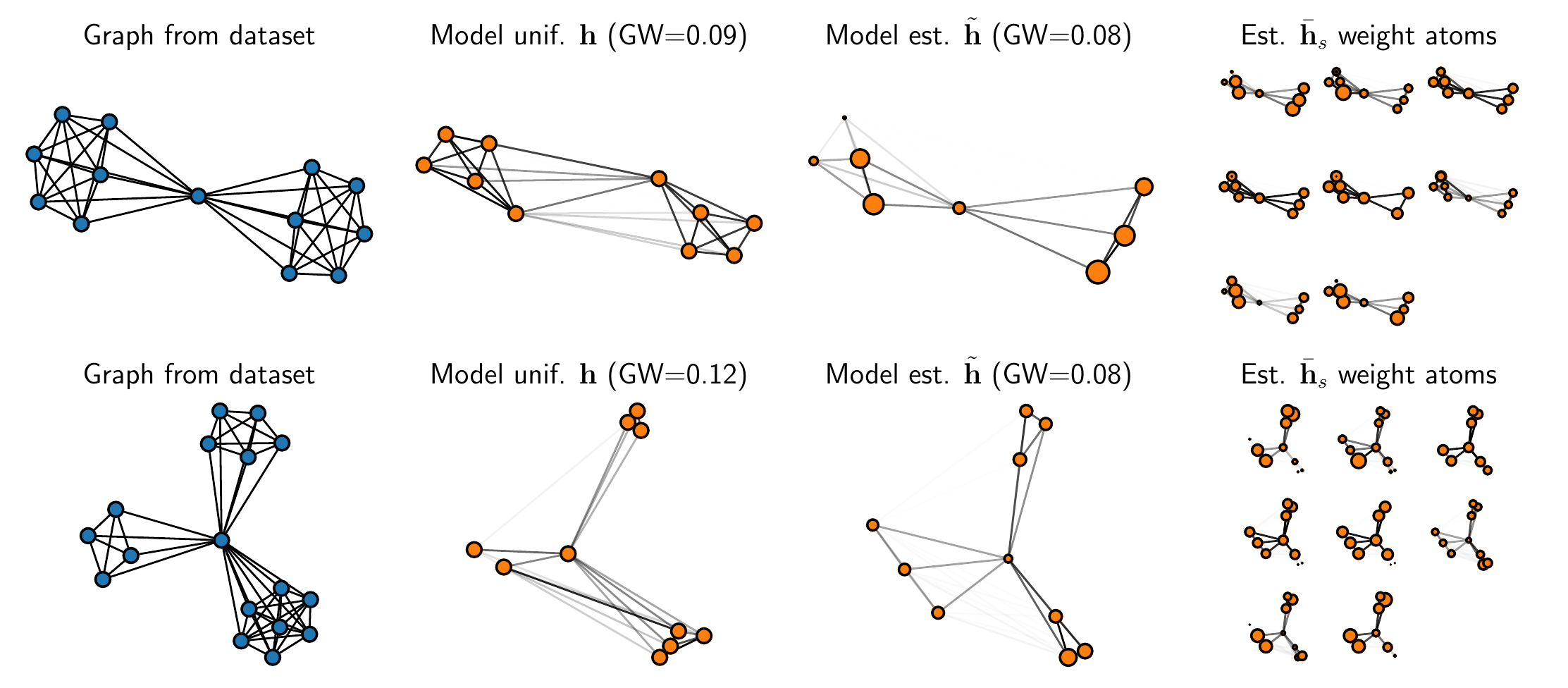}\vspace{-1mm}
		\caption{Modeling of two real life graphs from IMDB-M with our GDL
			approaches with 8 atoms of order 10. (left) original graphs 	from the dataset, (center left) linear model for GDL with uniform
			weights as in \eqref{eq:dl}, (center right) linear model for GDL with estimated
			weights  as in \eqref{eq:dl_h} and (right) different $\overline{\vh}_s$ on the
			estimated structure. }  \label{fig:unmix_weights}
	\end{figure}

	We illustrate in Fig. \ref{fig:unmix_weights} the interest of the extension of
	GDL with estimated weights for IMDB-M dataset. We can see in the center-left
	part of the figure that, without estimating the weights, GDL can experience
	difficulties producing a model that preserves the global structure of the graph
	because of the uniform weights on the nodes. In opposition, simultaneously
	estimating the weights brings a more representative modeling (in the GW sense),
	as illustrated in the centred-right columns. The weights estimation can
	re-balance and even discard non relevant nodes, in the vein of attention mechanisms. We report in the supplementary material a companion study for clustering tasks which further supports our extension concerning the learning of node weights.
	
	%	\vspace{-1mm}
	\subsection{Online graph subspace estimation and change detection}\label{subsec:real2}
	
	Finally we provide experiments for online graph subspace estimation on simulated
	and real life datasets. We show that our approach can be used for subspace
	tracking of graphs as well as for change point detection of subspaces. 
	
	\paragraph{Datasets and experiments} In this section we considered two new large graph
	classification datasets: TWITCH-EGOS \citep{karateclub} containing social graphs without
	attributes belonging to 2 classes and TRIANGLES
	\citep{knyazev2019understanding} that is a simulated dataset of labeled graphs
	with 10 classes. Here we investigate how our approach fits to online data, \ie\ in the presence of a stream of graphs. The experiments are designed with
	different time segments where each segment streams graphs belonging to the same
	classes (or group of classes). The aim is to see if the method learns the current
	stream and detects or adapts to abrupt changes in the stream. For TWITCH-EGOS, we first streamed all graphs of a class (A), then graphs of the
	other class (B), both counting more than 60.000 graphs. All these graphs consist in a unique high-frequency (a hub structure) with sparse connections between non-central nodes (sparser for class B).  For TRIANGLES, the 
	stream follows the three
	groups A,B and C, with 10,000 graphs each, where  the labels  associated with each group are:
	$A=\{4,5,6,7\}$, $B= \{8,9,10\}$ and $C= \{1,2,3\}$.
	
	\begin{figure}[t]
		\centering
		\hspace{-2mm}\includegraphics[width=.99\linewidth]{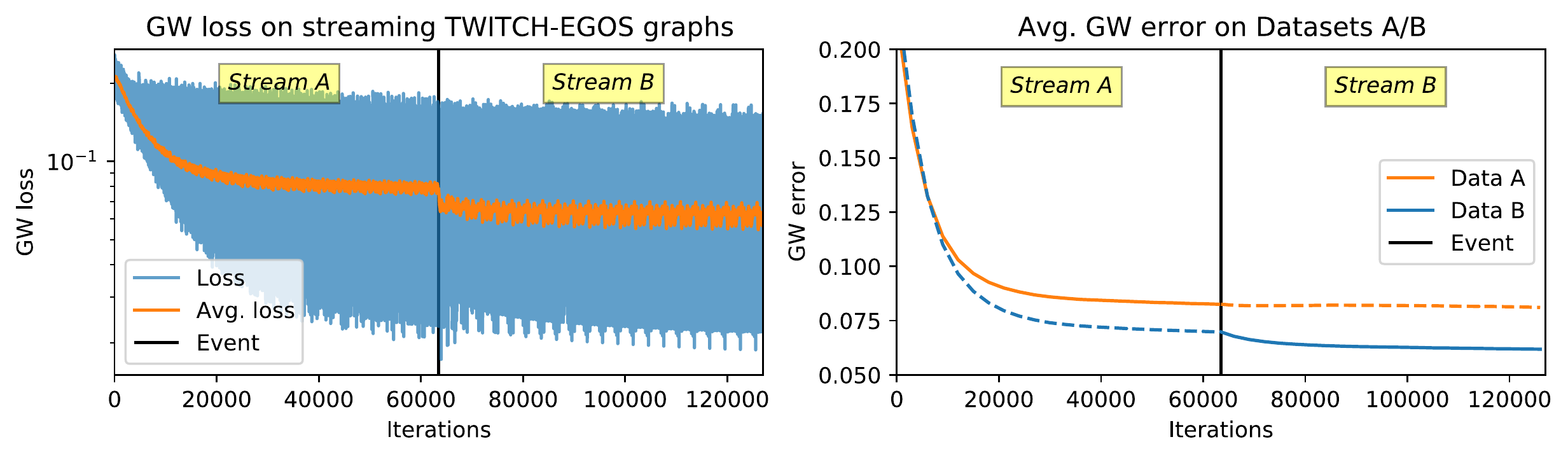}
		%\vspace{-4mm}
		\includegraphics[width=\linewidth]{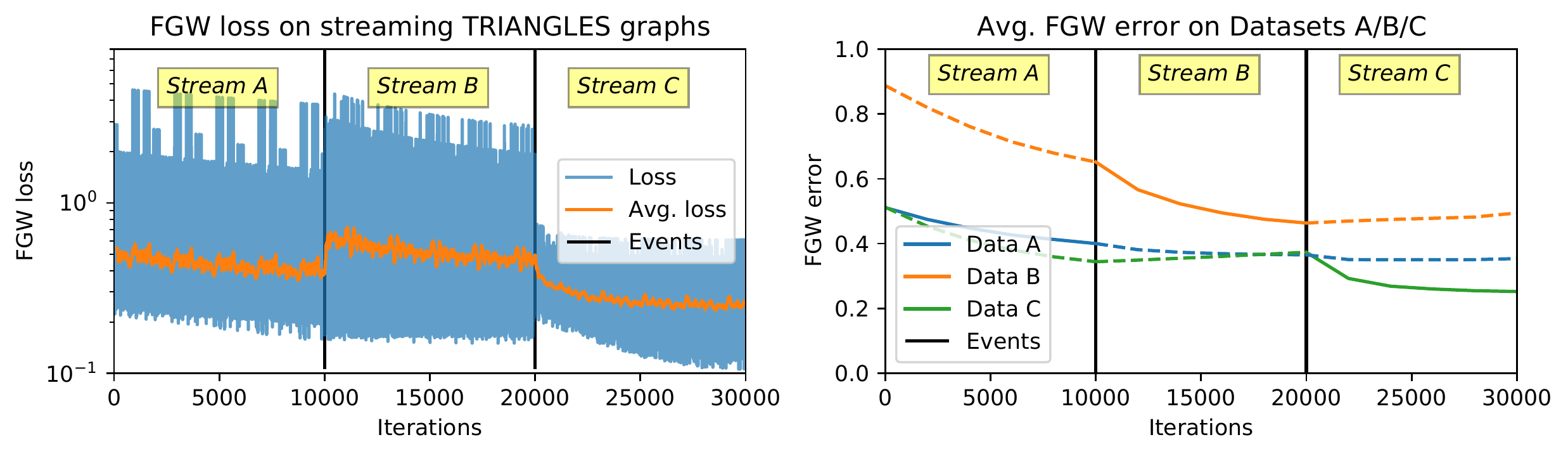}\vspace{-4mm}
		\caption{Online GDL on dataset TWITCH-EGOS with 2 atoms of
			14 nodes each (top) and on TRIANGLES with 4 atoms of 17 nodes each (bottom).} \label{fig:online}
	\end{figure}
	
	\paragraph{Results and discussion} The online (F)GW losses and a running
	mean of these losses are reported for
	each dataset on the left part of Fig.~\ref{fig:online}. One the right part of the Figure, we
	report the average losses computed on several datasets containing data from each stream at some
	time instant along the iterations. First, the online learning for both datasets
	can be seen in the running means with a clear decrease of loss on each time
	segment. Also,  note that at each event (change of
	stream) a jump in terms of loss is visible suggesting that the method can
	be used for change point detection. Finally it is interesting to see on the
	TRIANGLES dataset that while the loss on Data B is clearly decreased during
	Stream B it increases again during Stream C, thus showing that our algorithm performs subspace tracking, adapting to the new data and forgetting old subspaces no longer necessary.  
	
	\section{Conclusion}
	
	We present a new \emph{linear} Dictionary Learning approach for
	graphs with different orders relying on the Gromov Wasserstein (GW) divergence, where graphs are modeled as convex combination of graph atoms. We design an online stochastic algorithm to efficiently learn our dictionary and propose a computationally light proxy to the GW distance in the described graphs subspace. Our experiments on clustering classification and online subspace tracking demonstrate the interest of our unsupervised representation learning approach. We envision several extensions to this work, notably in the context of
	graph denoising or graph inpainting.
	
	\section*{Acknowledgments}
	This work is partially funded through the projects OATMIL ANR-17-CE23-0012, OTTOPIA ANR-20-CHIA-0030 
	and 3IA C\^{o}te d'Azur Investments ANR-19-P3IA-0002 of the French National Research
	Agency (ANR). This research was produced within the framework of Energy4Climate
	Interdisciplinary Center (E4C) of IP Paris and Ecole des Ponts ParisTech. This
	research was supported by 3rd Programme d'Investissements d'Avenir
	ANR-18-EUR-0006-02. This action benefited from the support of the Chair
	"Challenging Technology for Responsible Energy" led by l'X – Ecole polytechnique
	and the Fondation de l'Ecole polytechnique, sponsored by TOTAL. This work is supported by the ACADEMICS grant of the IDEXLYON, project of the Université de Lyon, PIA operated by ANR-16-IDEX-0005.
	The authors are grateful to the OPAL infrastructure from Universit\'{e} C\^{o}te d'Azur for providing resources and support.
	
	\bibliography{citations}

\begin{thebibliography}{75}
\providecommand{\natexlab}[1]{#1}
\providecommand{\url}[1]{\texttt{#1}}
\expandafter\ifx\csname urlstyle\endcsname\relax
  \providecommand{\doi}[1]{doi: #1}\else
  \providecommand{\doi}{doi: \begingroup \urlstyle{rm}\Url}\fi

\bibitem[Arjovsky et~al.(2017)Arjovsky, Chintala, and
  Bottou]{arjovsky2017wasserstein}
Arjovsky, M., Chintala, S., and Bottou, L.
\newblock Wasserstein gan.
\newblock \emph{arXiv preprint arXiv:1701.07875}, 2017.

\bibitem[Barbe et~al.(2020)Barbe, Sebban, Gon{\c c}alves, Borgnat, and
  Gribonval]{barbe:hal-02795056}
Barbe, A., Sebban, M., Gon{\c c}alves, P., Borgnat, P., and Gribonval, R.
\newblock {Graph Diffusion Wasserstein Distances}.
\newblock In \emph{{European Conference on Machine Learning and Principles and
  Practice of Knowledge Discovery in Databases}}, Ghent, Belgium, September
  2020.

\bibitem[Bavaud(2010)]{bavaud2010euclidean}
Bavaud, F.
\newblock Euclidean distances, soft and spectral clustering on weighted graphs.
\newblock In \emph{Joint European Conference on Machine Learning and Knowledge
  Discovery in Databases}, pp.\  103--118. Springer, 2010.

\bibitem[Bobadilla et~al.(2013)Bobadilla, Ortega, Hernando, and
  Guti{\'e}rrez]{bobadilla2013recommender}
Bobadilla, J., Ortega, F., Hernando, A., and Guti{\'e}rrez, A.
\newblock Recommender systems survey.
\newblock \emph{Knowledge-based systems}, 46:\penalty0 109--132, 2013.

\bibitem[Bonneel et~al.(2016)Bonneel, Peyr{\'e}, and Cuturi]{BTSSPP15}
Bonneel, N., Peyr{\'e}, G., and Cuturi, M.
\newblock Wasserstein barycentric coordinates: Histogram regression using
  optimal transport.
\newblock \emph{ACM Transactions on Graphics (Proceedings of SIGGRAPH 2016)},
  35\penalty0 (4), 2016.

\bibitem[Borgwardt \& Kriegel(2005)Borgwardt and
  Kriegel]{borgwardt2005shortest}
Borgwardt, K.~M. and Kriegel, H.-P.
\newblock Shortest-path kernels on graphs.
\newblock In \emph{Fifth IEEE international conference on data mining
  (ICDM'05)}, pp.\  8--pp. IEEE, 2005.

\bibitem[Bronstein et~al.(2017)Bronstein, Bruna, LeCun, Szlam, and
  Vandergheynst]{bronstein2017geometric}
Bronstein, M.~M., Bruna, J., LeCun, Y., Szlam, A., and Vandergheynst, P.
\newblock Geometric deep learning: going beyond euclidean data.
\newblock \emph{IEEE Signal Processing Magazine}, 34\penalty0 (4):\penalty0
  18--42, 2017.

\bibitem[Cand{\`e}s et~al.(2011)Cand{\`e}s, Li, Ma, and
  Wright]{candes-robust-2009}
Cand{\`e}s, E.~J., Li, X., Ma, Y., and Wright, J.
\newblock Robust principal component analysis?
\newblock \emph{Journal of the ACM (JACM)}, 58\penalty0 (3):\penalty0 1--37,
  2011.

\bibitem[Chen et~al.(2001)Chen, Donoho, and Saunders]{chen2001atomic}
Chen, S.~S., Donoho, D.~L., and Saunders, M.~A.
\newblock Atomic decomposition by basis pursuit.
\newblock \emph{SIAM review}, 43\penalty0 (1):\penalty0 129--159, 2001.

\bibitem[Chowdhury \& M{\'e}moli(2019)Chowdhury and
  M{\'e}moli]{chowdhury2019gromov}
Chowdhury, S. and M{\'e}moli, F.
\newblock The gromov--wasserstein distance between networks and stable network
  invariants.
\newblock \emph{Information and Inference: A Journal of the IMA}, 8\penalty0
  (4):\penalty0 757--787, 2019.

\bibitem[Chowdhury \& Needham(2020)Chowdhury and
  Needham]{chowdhury-generalized-2020}
Chowdhury, S. and Needham, T.
\newblock Generalized {Spectral} {Clustering} via {Gromov}-{Wasserstein}
  {Learning}.
\newblock \emph{arXiv:2006.04163 [cs, math, stat]}, June 2020.
\newblock arXiv: 2006.04163.

\bibitem[Cuturi \& Blondel(2018)Cuturi and Blondel]{cuturi-soft-dtw-2018}
Cuturi, M. and Blondel, M.
\newblock Soft-{DTW}: a {Differentiable} {Loss} {Function} for {Time}-{Series}.
\newblock \emph{arXiv:1703.01541 [stat]}, February 2018.
\newblock arXiv: 1703.01541.

\bibitem[Day(1985)]{day-optimal-1985}
Day, W.~H.
\newblock Optimal algorithms for comparing trees with labeled leaves.
\newblock \emph{Journal of classification}, 2\penalty0 (1):\penalty0 7--28,
  1985.

\bibitem[Debnath et~al.(1991)Debnath, Lopez~de Compadre, Debnath, Shusterman,
  and Hansch]{debnath1991structure}
Debnath, A.~K., Lopez~de Compadre, R.~L., Debnath, G., Shusterman, A.~J., and
  Hansch, C.
\newblock Structure-activity relationship of mutagenic aromatic and
  heteroaromatic nitro compounds. correlation with molecular orbital energies
  and hydrophobicity.
\newblock \emph{Journal of medicinal chemistry}, 34\penalty0 (2):\penalty0
  786--797, 1991.

\bibitem[Ditzler et~al.(2015)Ditzler, Roveri, Alippi, and Polikar]{Ditzler2015}
Ditzler, G., Roveri, M., Alippi, C., and Polikar, R.
\newblock Learning in nonstationary environments: A survey.
\newblock \emph{Computational Intelligence Magazine, IEEE}, 10:\penalty0
  12--25, 11 2015.

\bibitem[Feragen et~al.(2013)Feragen, Kasenburg, Petersen, de~Bruijne, and
  Borgwardt]{feragen2013scalable}
Feragen, A., Kasenburg, N., Petersen, J., de~Bruijne, M., and Borgwardt, K.
\newblock Scalable kernels for graphs with continuous attributes.
\newblock In \emph{Advances in neural information processing systems}, pp.\
  216--224, 2013.

\bibitem[Flamary \& Courty(2017)Flamary and Courty]{flamary2017pot}
Flamary, R. and Courty, N.
\newblock Pot python optimal transport library.
\newblock \emph{GitHub: https://github. com/rflamary/POT}, 2017.

\bibitem[G{\"a}rtner et~al.(2003)G{\"a}rtner, Flach, and
  Wrobel]{gartner2003graph}
G{\"a}rtner, T., Flach, P., and Wrobel, S.
\newblock On graph kernels: Hardness results and efficient alternatives.
\newblock In \emph{Learning theory and kernel machines}, pp.\  129--143.
  Springer, 2003.

\bibitem[Grattarola et~al.(2019)Grattarola, Zambon, Livi, and
  Alippi]{Grattarola}
Grattarola, D., Zambon, D., Livi, L., and Alippi, C.
\newblock Change detection in graph streams by learning graph embeddings on
  constant-curvature manifolds.
\newblock \emph{IEEE Transactions on Neural Networks and Learning Systems},
  PP:\penalty0 1--14, 07 2019.

\bibitem[Harchaoui \& Bach(2007)Harchaoui and Bach]{harchaoui-image-2007}
Harchaoui, Z. and Bach, F.
\newblock Image classification with segmentation graph kernels.
\newblock In \emph{2007 IEEE Conference on Computer Vision and Pattern
  Recognition}, pp.\  1--8. IEEE, 2007.

\bibitem[Heitmann \& Breakspear(2018)Heitmann and Breakspear]{Heitmann2018}
Heitmann, S. and Breakspear, M.
\newblock Putting the "dynamic" back into dynamic functional connectivity.
\newblock \emph{Network Neuroscience}, 2\penalty0 (2):\penalty0 150--174, 2018.

\bibitem[Holland et~al.(1983)Holland, Laskey, and
  Leinhardt]{holland1983stochastic}
Holland, P.~W., Laskey, K.~B., and Leinhardt, S.
\newblock Stochastic blockmodels: First steps.
\newblock \emph{Social networks}, 5\penalty0 (2):\penalty0 109--137, 1983.

\bibitem[Ioffe \& Szegedy(2015)Ioffe and Szegedy]{ioffe2015batch}
Ioffe, S. and Szegedy, C.
\newblock Batch normalization: Accelerating deep network training by reducing
  internal covariate shift.
\newblock In \emph{International conference on machine learning}, pp.\
  448--456. PMLR, 2015.

\bibitem[Jaggi(2013)]{jaggi2013revisiting}
Jaggi, M.
\newblock Revisiting frank-wolfe: Projection-free sparse convex optimization.
\newblock In \emph{International Conference on Machine Learning}, pp.\
  427--435. PMLR, 2013.

\bibitem[Kingma \& Ba(2014)Kingma and Ba]{kingma2014adam}
Kingma, D.~P. and Ba, J.
\newblock Adam: A method for stochastic optimization.
\newblock \emph{arXiv preprint arXiv:1412.6980}, 2014.

\bibitem[Knyazev et~al.(2019)Knyazev, Taylor, and
  Amer]{knyazev2019understanding}
Knyazev, B., Taylor, G.~W., and Amer, M.~R.
\newblock Understanding attention and generalization in graph neural networks.
\newblock \emph{arXiv preprint arXiv:1905.02850}, 2019.

\bibitem[Krichene et~al.(2015)Krichene, Krichene, and
  Bayen]{krichene2015efficient}
Krichene, W., Krichene, S., and Bayen, A.
\newblock Efficient bregman projections onto the simplex.
\newblock In \emph{2015 54th IEEE Conference on Decision and Control (CDC)},
  pp.\  3291--3298. IEEE, 2015.

\bibitem[Kriege et~al.(2018)Kriege, Fey, Fisseler, Mutzel, and
  Weichert]{kriege-recognizing-2018}
Kriege, N.~M., Fey, M., Fisseler, D., Mutzel, P., and Weichert, F.
\newblock Recognizing {Cuneiform} {Signs} {Using} {Graph} {Based} {Methods}.
\newblock \emph{arXiv:1802.05908 [cs]}, March 2018.
\newblock arXiv: 1802.05908.

\bibitem[Ktena et~al.(2017)Ktena, Parisot, Ferrante, Rajchl, Lee, Glocker, and
  Rueckert]{ktena-distance-2017}
Ktena, S.~I., Parisot, S., Ferrante, E., Rajchl, M., Lee, M., Glocker, B., and
  Rueckert, D.
\newblock Distance metric learning using graph convolutional networks:
  Application to functional brain networks.
\newblock In \emph{International Conference on Medical Image Computing and
  Computer-Assisted Intervention}, pp.\  469--477. Springer, 2017.

\bibitem[Lacoste-Julien(2016)]{lacoste-julien-convergence-2016}
Lacoste-Julien, S.
\newblock Convergence rate of frank-wolfe for non-convex objectives.
\newblock \emph{arXiv preprint arXiv:1607.00345}, 2016.

\bibitem[Li et~al.(2016)Li, Rangapuram, and Slawski]{li2016methods}
Li, P., Rangapuram, S.~S., and Slawski, M.
\newblock Methods for sparse and low-rank recovery under simplex constraints.
\newblock \emph{arXiv preprint arXiv:1605.00507}, 2016.

\bibitem[Mairal et~al.(2009)Mairal, Bach, Ponce, and Sapiro]{mairal2009online}
Mairal, J., Bach, F., Ponce, J., and Sapiro, G.
\newblock Online dictionary learning for sparse coding.
\newblock In \emph{Proceedings of the 26th annual international conference on
  machine learning}, pp.\  689--696, 2009.

\bibitem[Maretic et~al.(2019)Maretic, El~Gheche, Chierchia, and
  Frossard]{maretic2019got}
Maretic, H.~P., El~Gheche, M., Chierchia, G., and Frossard, P.
\newblock Got: an optimal transport framework for graph comparison.
\newblock In \emph{Advances in Neural Information Processing Systems}, pp.\
  13876--13887, 2019.

\bibitem[Masuda \& Lambiotte(2020)Masuda and Lambiotte]{masuda2020guide}
Masuda, N. and Lambiotte, R.
\newblock \emph{A Guide To Temporal Networks}, volume~6.
\newblock World Scientific, 2020.

\bibitem[M{\'e}moli(2011)]{memoli-gromovwasserstein-2011}
M{\'e}moli, F.
\newblock Gromov--wasserstein distances and the metric approach to object
  matching.
\newblock \emph{Foundations of computational mathematics}, 11\penalty0
  (4):\penalty0 417--487, 2011.

\bibitem[Murty(1988)]{murty-linear-1988}
Murty, K.
\newblock \emph{Linear {Complementarity}, {Linear} and {Nonlinear}
  {Programming}}.
\newblock Sigma series in applied mathematics. Heldermann, 1988.
\newblock ISBN 978-3-88538-403-8.

\bibitem[Narayanamurthy \& Vaswani(2018)Narayanamurthy and
  Vaswani]{narayanamurthy2018nearly}
Narayanamurthy, P. and Vaswani, N.
\newblock Nearly optimal robust subspace tracking.
\newblock In \emph{International Conference on Machine Learning}, pp.\
  3701--3709. PMLR, 2018.

\bibitem[Neumann et~al.(2016)Neumann, Garnett, Bauckhage, and
  Kersting]{neumann2016propagation}
Neumann, M., Garnett, R., Bauckhage, C., and Kersting, K.
\newblock Propagation kernels: efficient graph kernels from propagated
  information.
\newblock \emph{Machine Learning}, 102\penalty0 (2):\penalty0 209--245, 2016.

\bibitem[Ng et~al.(2002)Ng, Jordan, Weiss, et~al.]{ng-spectral-2002}
Ng, A.~Y., Jordan, M.~I., Weiss, Y., et~al.
\newblock On spectral clustering: Analysis and an algorithm.
\newblock \emph{Advances in neural information processing systems}, 2:\penalty0
  849--856, 2002.

\bibitem[Niepert et~al.(2016)Niepert, Ahmed, and
  Kutzkov]{niepert-learning-2016}
Niepert, M., Ahmed, M., and Kutzkov, K.
\newblock Learning convolutional neural networks for graphs.
\newblock In \emph{International conference on machine learning}, pp.\
  2014--2023, 2016.

\bibitem[Nikolentzos et~al.(2017)Nikolentzos, Meladianos, and
  Vazirgiannis]{DBLP:conf/aaai/NikolentzosMV17}
Nikolentzos, G., Meladianos, P., and Vazirgiannis, M.
\newblock Matching node embeddings for graph similarity.
\newblock In \emph{Proceedings of the Thirty-First {AAAI} Conference on
  Artificial Intelligence, February 4-9, 2017, San Francisco, California,
  {USA.}}, pp.\  2429--2435, 2017.

\bibitem[Perozzi et~al.(2014)Perozzi, Al-Rfou, and Skiena]{perozzi2014deepwalk}
Perozzi, B., Al-Rfou, R., and Skiena, S.
\newblock Deepwalk: Online learning of social representations.
\newblock In \emph{Proceedings of the 20th ACM SIGKDD international conference
  on Knowledge discovery and data mining}, pp.\  701--710, 2014.

\bibitem[Peyr\'{e} \& Cuturi(2019)Peyr\'{e} and
  Cuturi]{peyre-computational-2020}
Peyr\'{e}, G. and Cuturi, M.
\newblock Computational optimal transport.
\newblock \emph{Foundations and Trends in Machine Learning}, 11:\penalty0
  355--607, 2019.

\bibitem[Peyr{\'e} et~al.(2016)Peyr{\'e}, Cuturi, and Solomon]{peyre2016gromov}
Peyr{\'e}, G., Cuturi, M., and Solomon, J.
\newblock Gromov-wasserstein averaging of kernel and distance matrices.
\newblock In \emph{International Conference on Machine Learning}, pp.\
  2664--2672, 2016.

\bibitem[Rand(1971)]{rand1971objective}
Rand, W.~M.
\newblock Objective criteria for the evaluation of clustering methods.
\newblock \emph{Journal of the American Statistical association}, 66\penalty0
  (336):\penalty0 846--850, 1971.

\bibitem[Rolet et~al.(2016)Rolet, Cuturi, and Peyr{\'e}]{rolet2016fast}
Rolet, A., Cuturi, M., and Peyr{\'e}, G.
\newblock Fast dictionary learning with a smoothed wasserstein loss.
\newblock In \emph{Artificial Intelligence and Statistics}, pp.\  630--638.
  PMLR, 2016.

\bibitem[Rozemberczki et~al.(2020)Rozemberczki, Kiss, and Sarkar]{karateclub}
Rozemberczki, B., Kiss, O., and Sarkar, R.
\newblock {Karate Club: An API Oriented Open-source Python Framework for
  Unsupervised Learning on Graphs}.
\newblock In \emph{Proceedings of the 29th ACM International Conference on
  Information and Knowledge Management (CIKM '20)}, pp.\  3125–3132. ACM,
  2020.

\bibitem[Scarselli et~al.(2008)Scarselli, Gori, Tsoi, Hagenbuchner, and
  Monfardini]{scarselli2008graph}
Scarselli, F., Gori, M., Tsoi, A.~C., Hagenbuchner, M., and Monfardini, G.
\newblock The graph neural network model.
\newblock \emph{IEEE Transactions on Neural Networks}, 20\penalty0
  (1):\penalty0 61--80, 2008.

\bibitem[Schmitz et~al.(2018)Schmitz, Heitz, Bonneel, Ngole, Coeurjolly,
  Cuturi, Peyr{\'e}, and Starck]{schmitz2018wasserstein}
Schmitz, M.~A., Heitz, M., Bonneel, N., Ngole, F., Coeurjolly, D., Cuturi, M.,
  Peyr{\'e}, G., and Starck, J.-L.
\newblock Wasserstein dictionary learning: Optimal transport-based unsupervised
  nonlinear dictionary learning.
\newblock \emph{SIAM Journal on Imaging Sciences}, 11\penalty0 (1):\penalty0
  643--678, 2018.

\bibitem[Shervashidze et~al.(2009)Shervashidze, Vishwanathan, Petri, Mehlhorn,
  and Borgwardt]{shervashidze09}
Shervashidze, N., Vishwanathan, S., Petri, T., Mehlhorn, K., and Borgwardt, K.
\newblock Efficient graphlet kernels for large graph comparison.
\newblock In \emph{Artificial intelligence and statistics}, pp.\  488--495.
  PMLR, 2009.

\bibitem[Shi \& Malik(2000)Shi and Malik]{shi2000normalized}
Shi, J. and Malik, J.
\newblock Normalized cuts and image segmentation.
\newblock \emph{IEEE Transactions on pattern analysis and machine
  intelligence}, 22\penalty0 (8):\penalty0 888--905, 2000.

\bibitem[Siglidis et~al.(2020)Siglidis, Nikolentzos, Limnios, Giatsidis,
  Skianis, and Vazirgiannis]{siglidis2020grakel}
Siglidis, G., Nikolentzos, G., Limnios, S., Giatsidis, C., Skianis, K., and
  Vazirgiannis, M.
\newblock Grakel: A graph kernel library in python.
\newblock \emph{Journal of Machine Learning Research}, 21\penalty0
  (54):\penalty0 1--5, 2020.

\bibitem[Solomon et~al.(2016)Solomon, Peyr{\'e}, Kim, and
  Sra]{solomon_entropic_2016}
Solomon, J., Peyr{\'e}, G., Kim, V.~G., and Sra, S.
\newblock Entropic metric alignment for correspondence problems.
\newblock \emph{ACM Transactions on Graphics (TOG)}, 35\penalty0 (4):\penalty0
  1--13, 2016.

\bibitem[Srivastava et~al.(2014)Srivastava, Hinton, Krizhevsky, Sutskever, and
  Salakhutdinov]{srivastava2014dropout}
Srivastava, N., Hinton, G., Krizhevsky, A., Sutskever, I., and Salakhutdinov,
  R.
\newblock Dropout: a simple way to prevent neural networks from overfitting.
\newblock \emph{The journal of machine learning research}, 15\penalty0
  (1):\penalty0 1929--1958, 2014.

\bibitem[Stella \& Shi(2003)Stella and Shi]{stella2003multiclass}
Stella, X.~Y. and Shi, J.
\newblock Multiclass spectral clustering.
\newblock In \emph{null}, pp.\  313. IEEE, 2003.

\bibitem[Sturm(2012)]{sturm2012space}
Sturm, K.-T.
\newblock The space of spaces: curvature bounds and gradient flows on the space
  of metric measure spaces.
\newblock \emph{arXiv preprint arXiv:1208.0434}, 2012.

\bibitem[Sutherland et~al.(2003)Sutherland, O'brien, and
  Weaver]{sutherland2003spline}
Sutherland, J.~J., O'brien, L.~A., and Weaver, D.~F.
\newblock Spline-fitting with a genetic algorithm: A method for developing
  classification structure- activity relationships.
\newblock \emph{Journal of chemical information and computer sciences},
  43\penalty0 (6):\penalty0 1906--1915, 2003.

\bibitem[Togninalli et~al.(2019)Togninalli, Ghisu, Llinares-L{\'o}pez, Rieck,
  and Borgwardt]{Togninalli19}
Togninalli, M., Ghisu, E., Llinares-L{\'o}pez, F., Rieck, B., and Borgwardt, K.
\newblock Wasserstein weisfeiler--lehman graph kernels.
\newblock In Wallach, H., Larochelle, H., Beygelzimer, A., d'Alch\'{e}{-}Buc,
  F., Fox, E., and Garnett, R. (eds.), \emph{Advances in Neural Information
  Processing Systems~32~(NeurIPS)}, pp.\  6436--6446. Curran Associates, Inc.,
  2019.

\bibitem[Tseng(2001)]{tseng2001convergence}
Tseng, P.
\newblock Convergence of a block coordinate descent method for
  nondifferentiable minimization.
\newblock \emph{Journal of optimization theory and applications}, 109\penalty0
  (3):\penalty0 475--494, 2001.

\bibitem[Vayer et~al.(2018)Vayer, Chapel, Flamary, Tavenard, and
  Courty]{vayer-fused-2018}
Vayer, T., Chapel, L., Flamary, R., Tavenard, R., and Courty, N.
\newblock Fused gromov-wasserstein distance for structured objects: theoretical
  foundations and mathematical properties.
\newblock \emph{arXiv preprint arXiv:1811.02834}, 2018.

\bibitem[Vayer et~al.(2019)Vayer, Courty, Tavenard, and
  Flamary]{vayer-optimal-nodate}
Vayer, T., Courty, N., Tavenard, R., and Flamary, R.
\newblock Optimal transport for structured data with application on graphs.
\newblock In \emph{International Conference on Machine Learning}, pp.\
  6275--6284. PMLR, 2019.

\bibitem[Villani(2003)]{villani-topics-2003}
Villani, C.
\newblock \emph{Topics in optimal transportation}.
\newblock Number~58. American Mathematical Soc., 2003.

\bibitem[Vishwanathan et~al.(2010)Vishwanathan, Schraudolph, Kondor, and
  Borgwardt]{vishwanathan2010graph}
Vishwanathan, S. V.~N., Schraudolph, N.~N., Kondor, R., and Borgwardt, K.~M.
\newblock Graph kernels.
\newblock \emph{The Journal of Machine Learning Research}, 11:\penalty0
  1201--1242, 2010.

\bibitem[Vlaski et~al.(2018)Vlaski, Mareti{\'c}, Nassif, Frossard, and
  Sayed]{vlaski2018online}
Vlaski, S., Mareti{\'c}, H.~P., Nassif, R., Frossard, P., and Sayed, A.~H.
\newblock Online graph learning from sequential data.
\newblock In \emph{2018 IEEE Data Science Workshop (DSW)}, pp.\  190--194.
  IEEE, 2018.

\bibitem[Wang et~al.(2020)Wang, Song, Wu, and Wang]{Wang2020StreamingGN}
Wang, J., Song, G., Wu, Y., and Wang, L.
\newblock Streaming graph neural networks via continual learning.
\newblock \emph{Proceedings of the 29th ACM International Conference on
  Information \& Knowledge Management}, 2020.

\bibitem[Wang \& Wong(1987)Wang and Wong]{wang1987stochastic}
Wang, Y.~J. and Wong, G.~Y.
\newblock Stochastic blockmodels for directed graphs.
\newblock \emph{Journal of the American Statistical Association}, 82\penalty0
  (397):\penalty0 8--19, 1987.

\bibitem[Wu et~al.(2020)Wu, Pan, Chen, Long, Zhang, and
  Philip]{wu2020comprehensive}
Wu, Z., Pan, S., Chen, F., Long, G., Zhang, C., and Philip, S.~Y.
\newblock A comprehensive survey on graph neural networks.
\newblock \emph{IEEE Transactions on Neural Networks and Learning Systems},
  2020.

\bibitem[Xu(2020)]{xu_gromov-wasserstein_2019}
Xu, H.
\newblock Gromov-wasserstein factorization models for graph clustering.
\newblock In \emph{Proceedings of the AAAI Conference on Artificial
  Intelligence}, volume~34, pp.\  6478--6485, 2020.

\bibitem[Xu et~al.(2019{\natexlab{a}})Xu, Luo, and Carin]{xu2019scalable}
Xu, H., Luo, D., and Carin, L.
\newblock Scalable gromov-wasserstein learning for graph partitioning and
  matching.
\newblock \emph{arXiv preprint arXiv:1905.07645}, 2019{\natexlab{a}}.

\bibitem[Xu et~al.(2019{\natexlab{b}})Xu, Luo, Zha, and Duke]{xu2019gromov}
Xu, H., Luo, D., Zha, H., and Duke, L.~C.
\newblock Gromov-wasserstein learning for graph matching and node embedding.
\newblock In \emph{International conference on machine learning}, pp.\
  6932--6941. PMLR, 2019{\natexlab{b}}.

\bibitem[Xu et~al.(2018)Xu, Hu, Leskovec, and Jegelka]{xu2018powerful}
Xu, K., Hu, W., Leskovec, J., and Jegelka, S.
\newblock How powerful are graph neural networks?
\newblock \emph{arXiv preprint arXiv:1810.00826}, 2018.

\bibitem[Yanardag \& Vishwanathan(2015)Yanardag and
  Vishwanathan]{yanardag-deep-2015}
Yanardag, P. and Vishwanathan, S.
\newblock Deep graph kernels.
\newblock In \emph{Proceedings of the 21th ACM SIGKDD international conference
  on knowledge discovery and data mining}, pp.\  1365--1374, 2015.

\bibitem[Yang et~al.(2018)Yang, Zhao, and Gao]{yang2018bandit}
Yang, P., Zhao, P., and Gao, X.
\newblock Bandit online learning on graphs via adaptive optimization.
\newblock International Joint Conferences on Artificial Intelligence, 2018.

\bibitem[Zambon et~al.(2017)Zambon, Alippi, and Livi]{Zambon}
Zambon, D., Alippi, C., and Livi, L.
\newblock Concept drift and anomaly detection in graph streams.
\newblock \emph{IEEE Transactions on Neural Networks and Learning Systems}, PP,
  06 2017.

\bibitem[Zambon et~al.(2019)Zambon, Alippi, and Livi]{Zambon2019ChangePointMO}
Zambon, D., Alippi, C., and Livi, L.
\newblock Change-point methods on a sequence of graphs.
\newblock \emph{IEEE Transactions on Signal Processing}, 67:\penalty0
  6327--6341, 2019.

\end{thebibliography}
	\bibliographystyle{icml2021}
	\newpage
	
	\onecolumn
	\section{Supplementary Material}
	\subsection{Notations \& definitions}\label{sec:defs}
	In this section we recall the notations used in the rest of the supplementary. 
	
	For matrices we note  $S_N(\R)$ the set of symmetric matrices in $\R^{N\times N}$ and $\scalar{\cdot}{\cdot}_F$ the Frobenius inner product defined for real matrices $\mC_1,\mC_2$ as $\scalar{\mC_1}{\mC_2}_F=\tr(\mC_1^{\top}\mC_2)$ where $\tr$ denotes the trace of matrices. Moreover $\mC_1\odot\mC_2$ denotes the Hadamard product of $\mC_1,\mC_2$, \ie\ $(\mC_1\odot\mC_2)_{ij}=C_1(i,j)C_2(i,j)$. Finally $\vec(\mC)$ denotes the vectorization of the matrix $\mC$.
	
	For vectors the Euclidean norm is denoted as $\|\cdot\|_2$ associated with the inner product $\scalar{\cdot}{\cdot}$. For a vector $\xbf \in \R^{N}$ the operator $\diag(\xbf)$ denotes the diagonal matrix defined with the values of $\xbf$. If $\Mbf \in S_N(\R)$ is a positive semi-definite matrix we note $\|\cdot\|_{\Mbf}$ the pseudo-norm defined for $\xbf\in \R^{N}$ by $\|\xbf\|^{2}_{\Mbf}=\xbf^{\top}\Mbf\xbf$. By some abuse of terminology we will use the term Mahalanobis distance to refer to generalized quadratic distances defined as $d_{\Mbf}(\xbf,\ybf)=\|\xbf-\ybf\|_{\Mbf}$. The fact that $\Mbf$ is positive semi-definite ensures that $d_{\Mbf}$ satisfies the
	properties of a pseudo-distance. 
	
	For a $4$-D tensor $\L=(L_{ijkl})_{ijkl}$ we note $\otimes$ the tensor-matrix multiplication, \ie\ given a matrix $\mC$, $\L \otimes \Abf$ is the matrix $\left(\sum_{k,l} L_{i,j,k,l}A_{k,l}\right)_{i,j}$.
	
	The simplex of histograms (or \emph{weights}) with $N$ bins is $\Sigma_N := \left\{\mathbf{h}\in \mathbb{R}^+_N| \sum_i h_i = 1 \right\}$. For two histograms $\vh^X \in \Sigma_{N_X},\vh^Y \in \Sigma_{N_Y}$ the set $\mathcal{U}(\vh^X,\vh^Y):= \{\mT \in \R_{+}^{N^X\times N^Y}|\mT\mathbf{1}_{N^{Y}}=\vh^X, \mT^T\mathbf{1}_{N^{X}}=\vh^Y\}$ is the set of couplings between $\vh^X,\vh^Y$. 
	
	Recall that for two graphs $G^X=(\mC^{X},\h^{X})$ and $G^Y=(\mC^{Y},\h^{Y})$ the $GW_2$ distance between $G^X$ and $G^Y$ is defined as the result of the following optimization problem:
	\begin{equation}
	\label{eq:gwdef_supp}
	\min_{\mT \in \mathcal{U}(\vh^X,\vh^Y)} \sum_{ijkl} \left( C^X_{ij}-C^Y_{kl}\right)^2\ T_{ik}T_{jl} 
	\end{equation}
	In the following we denote by $GW_2(\mC^{X},\mC^{Y},\vh^X,\vh^Y)$ the optimal value of \eqref{eq:gwdef_supp} or  by $GW_2(\mC^{X},\mC^{Y})$ when the weights are uniform. With more compact notations:
	\begin{equation}
	GW_2(\mC^{X},\mC^{Y},\vh^X,\vh^Y)=\min_{\mT \in \mathcal{U}(\vh^X,\vh^Y)} \langle \L(\mC^{X},\mC^{Y})\otimes \mT,\mT\rangle_{F}
	\end{equation}
	where $\L(\mC^{X},\mC^{Y})$ is the $4$-D tensor $\L(\mC^{X},\mC^{Y})=\left((C^X_{ij}-C^Y_{kl})^2\right)_{ijkl}$
	
	For graphs with attributes we use the Fused Gromov-Wasserstein distance \cite{vayer-optimal-nodate}. More precisely consider two graphs $G^X=(\mC^{X},\mA^{X},\h^{X})$ and $G^Y=(\mC^{Y},\mA^{Y},\h^{Y})$ where $\mA^X=(\a_i^{X})_{i \in [N^X]} \in \R^{N^X \times
		d},\mA^Y=(\a_j^{Y})_{j \in [N^Y]} \in \R^{N^Y \times d}$ are the matrices of all features. Given $\alpha \in [0,1]$ and a cost function $c: \R^{d} \times \R^{d} \rightarrow \R$ between vectors in $\R^{d}$ the $\fgw_2$ distance is defined as the result of the following optimization problem:
	\begin{equation}
	\label{eq:fgwdef_supp}
	\min_{\mT \in \mathcal{U}(\vh^X,\vh^Y)} (1-\alpha) \sum_{ij}c(\a^{X}_i,\a_j^{Y})T_{ij}+\alpha\sum_{ijkl} \left( C^X_{ij}-C^Y_{kl}\right)^2 T_{ik}T_{jl} 
	\end{equation}
	In the following we note $\fgw_{2,\alpha}(\mC^{X},\mA^{X},\mC^{Y},\mA^{Y},\vh^X,\vh^Y)$ the optimal value of \eqref{eq:fgwdef_supp} or by $\fgw_{2,\alpha}(\mC^{X},\mA^{X},\mC^{Y},\mA^{Y})$ when the weights are uniform. The term $\sum_{ij}c(\a^{X}_i,\a_j^{Y})T_{ij}$ will be called the \emph{Wasserstein objective} and denoted as $\mathcal{F}(\mA^{X},\mA^{Y},\mT)$ and the term $\sum_{ijkl} \left( C^X_{ij}-C^Y_{kl}\right)^2 T_{ik}T_{jl}$ will be called the \emph{Gromov-Wasserstein objective} and denoted $\mathcal{E}(\mC^{X},\mC^{Y},\mT)$.
	
	\subsection{Proofs of the different results}
	\subsubsection{(F)GW upper-bounds in the embedding space}
	\begin{proposition}[Gromov-Wasserstein]
		\label{prop:embed_graph}
		For two embedded graphs with embeddings $\vw^{(1)}$ and $\vw^{(2)}$ over the set of pairwise relation matrices $\{\overline{\mC_s}\}_{s\in[S]} \subset S_N(\R)$, with a shared masses vector $\vh$, the following inequality holds
		\begin{equation}
		GW_2\left(\sum_{s \in [S]} w^{(1)}_s \overline{\mC_s}, \sum_{s \in [S]} w^{(2)}_s \overline{\mC_s}\right) \leq \|\vw^{(1)} - \vw^{(2)}\|_\mM \label{eq:mah_gw}
		\end{equation}
		where $\Mbf = (\scalar{\mD_{\vh}\overline{\mC_p}}{ \overline{\mC_q}\mD_{\vh}}_{F})_{pq}$ and $\mD_{\vh}= diag(\vh)$. $\mM$ is a positive semi-definite matrix hence engenders a Mahalanobis distance between embeddings.
	\end{proposition}
	\paragraph{Proof.} Let consider the formulation of the GW distance as a Frobenius inner product (see \emph{e.g} \citep{peyre2016gromov}). Denoting $\mT$ the optimal transport plan between both embedded graph and the power operation over matrices applied at entries level, 
	\begin{equation}\label{eq:rawGW_embedded}
	GW_2(\sum_s w^{(1)}_s \overline{\mC_s}, \sum_s w^{(2)}_s \overline{\mC_s},\vh) = \scalar{(\sum_s w_s^{(1)}\overline{\mC_s})^2\vh \mathbf{1}_N^{\top} +\mathbf{1}_N\vh^{\top}(\sum_s w_s^{(2)}\overline{\mC_s}^{\top})^2 - 2(\sum_s w_s^{(1)}\overline{\mC_s}) \mT(\sum_s w_s^{(2)}\overline{\mC_s}^{\top}) }{\mT}_F
	\end{equation}
	Using the marginal constraints of GW problem, \emph{i.e} $\mT \in \mathcal{U}(\vh,\vh):= \{\mT \in \R_{+}^{N\times N}|\mT\mathbf{1}_{N}=\vh, \mT^T\mathbf{1}_{N}=\vh\}$, and the symmetry of matrices $\{\overline{\mC_s}\}$,\eqref{eq:rawGW_embedded} can be developed as follow,
	\begin{equation}\label{eq:GW_embedded1}
	GW_2(\sum_s w^{(1)}_s \overline{\mC_s}, \sum_s w^{(2)}_s \overline{\mC_s},\vh) =  \sum_{pq} \tr\left(w_p^{(1)}w_q^{(1)}(\overline{\mC_p}\odot \overline{\mC_q})\vh\vh^{\top} + w_p^{(2)}w_q^{(2)} (\overline{\mC_p}\odot \overline{\mC_q})\vh\vh^{\top} - 2 w_p^{(1)}w_q^{(2)} \overline{\mC_p}\mT \overline{\mC_q} \mT^{\top}\right)
	\end{equation}
	
	With the following property of the trace operator:
	\begin{equation}
	\tr\left((\mC_1\odot\mC_2)\vx\vx^{\top}\right) = \tr\left(\mC_1^{\top} \diag(\vx) \mC_2 \diag(\vx)\right)
	\end{equation}
	Denoting $\mD_{\vh} = \diag(\vh)$, \eqref{eq:GW_embedded1} can be expressed as:
	\begin{equation}
	\begin{split}
	&\gw_2(\sum_p w^{(1)}_p \overline{\mC_p}, \sum_q w^{(2)}_q \overline{\mC_q},\vh)=\sum_{pq} (w_p^{(1)}w_q^{(1)} +w_p^{(2)}w_q^{(2)})\scalar{\mD_{\vh} \overline{\mC_p}}{\overline{\mC_q} \mD_{\vh}}_{F}-2 w_p^{(1)}w_q^{(2)}\scalar{\mT^{\top} \overline{\mC_p}}{\overline{\mC_q}\mT^{\top}}_{F}
	\end{split}
	\end{equation}
	As $\mT \in \mathcal{U}(\vh,\vh)$ is a minimum of the GW objective, we can bound by above \eqref{eq:GW_embedded1} by evaluating the GW objective in $\mD_{\vh} \in \mathcal{U}(\vh,\vh)$, which is a sub-optimal admissible coupling.
	\begin{equation}
	\begin{split}
	\gw_2(\sum_p w^{(1)}_p \overline{\mC_p}, \sum_q w^{(2)}_q \overline{\mC_q},\vh) &\leq \sum_{pq} (w_p^{(1)}w_q^{(1)}+w_p^{(2)}w_q^{(2)}- 2w_p^{(1)} w_q^{(2)})\scalar{\mD_{\vh}\overline{\mC_p}}{\overline{\mC_q} \mD_{\vh}}_{F}\\
	&= {\vw^{(1)}}^T\mM\vw^{(1)} +{\vw^{(2)}}^{\top} \mM \vw^{(2)} - 2{\vw^{(1)}}^{\top} \mM \vw^{(2)}\\
	\end{split}
	\end{equation}
	with $\mM= (\scalar{\mD_{\vh} \overline{\mC_p}}{\overline{\mC_q} \mD_{\vh}}_{F})_{pq}$. It suffices to prove that the matrix $\mM$ is a PSD matrix to conclude that it defines a Mahalanobis distance over the set of embeddings $\vw$ which bounds by above the GW distance between corresponding embedded graphs.
	Let consider the following reformulation of an entry $M_{pq}$ as follow,
	\begin{equation}
	\scalar{\mD_{\vh} \overline{\mC_p}}{\overline{\mC_q}\mD_{\vh}} = \vec(\mB_p)^{\top}\vec(\mB_q)
	\end{equation}
	where $\forall n \in [S], \mB_n = \mD_{\vh}^{1/2}\overline{\mC_n}\mD_{\vh}^{1/2}$. Hence with  $\mB= (\mB_n)_n \subset \R^{N^2 \times S}$ , $\mM$ can be factorized as $\mB^T\mB$ and therefore is a PSD matrix.
	$\square$
	
	A similar result can be proven for the Fused Gromov-Wasserstein distance:
	\begin{proposition}[Fused Gromov-Wasserstein]
		For two embedded graphs with node attributes, with embeddings $\vw^{(1)}$ and $\vw^{(2)}$ over the set of pairwise relation matrices $\{(\overline{\mC_s},\overline{\mA_s})\}_{s\in[S]} \subset S_N(\R) \times \R^{N\times dd}$, and a shared masses vector $\vh$, the following inequality holds $\forall \alpha \in (0,1)$,
		\begin{equation}
		\begin{split}
		FGW_{2,\alpha}\left(\widetilde{\mC}(\vw^{(1)}), \widetilde{\mA}(\vw^{(1)}), \widetilde{\mC}(\vw^{(2)}), \widetilde{\mA}(\vw^{(2)})\right) \label{eq:mah_fgw}\leq \| \vw^{(1)} - \vw^{(2)}\|_{\alpha\mM_1 +(1-\alpha)\mM_2}
		\end{split}
		\end{equation}
		with,
		\begin{equation}
		\widetilde{\mC}(\vw)= \sum_s w_s \overline{\mC_s}  \quad \textit{and} \quad \widetilde{\mA}(\vw)= \sum_s w_s \overline{\mA_s}
		\end{equation}
		Where $\mM_1 = \left(\scalar{\mD_{\vh}\overline{\mC_p}}{ \overline{\mC_q}\mD_{\vh}}_F\right)_{pq}$ and $\mM_2=(\scalar{\mD_{\vh}^{1/2}\overline{\mA_p}}{\mD_{\vh}^{1/2}\overline{\mA_q}}_F)_{pq \in [S]}$, and $\mD_{\vh} = diag(\vh)$, are PSD matrices and therefore their linear combination being PSD engender Mahalanobis distances over the unmixing space.
	\end{proposition} 
	\paragraph{Proof.}
	Let consider the optimal transport plan $\mT \in \mathcal{U}(\vh,\vh)$ of the $FGW$ distance between both embedded structures.
	\begin{equation} \label{eq:FGW_bound1}
	\begin{split}
	FGW_{2,\alpha}^2\left(\widetilde{\mC}(\vw^{(1)}),\widetilde{\mA}(\vw^{(1)}),\widetilde{\mC}(\vw^{(2)}),\widetilde{\mA}(\vw^{(2)}),\vh\right)&= \alpha \mathcal{E}\left(\widetilde{\mC}(\vw^{(1)}), \widetilde{\mC}(\vw^{(2)}),\mT\right) +(1-\alpha)\mathcal{F}\left(\widetilde{\mA}(\vw^{(1)}),\widetilde{\mA}(\vw^{(2)}), \mT\right)
	\end{split}
	\end{equation}
	where $\mathcal{E}$ and $\mathcal{F}$ denotes respectively the Gromov-Wasserstein objective and the Wasserstein objective. 
	As a similar approach than for Proposition \ref{eq:mah_gw} can be used for the GW objective involved in \eqref{eq:FGW_bound1}, we will first highlight a suitable factorization of the Wasserstein objective $\mathcal{F}$. Note that for any feature matrices $\mA_1=(\va_{1,i})_{i\in[N]},\mA_2=(\va_{2,i})_{i\in[N]} \in \R^{N*d}$, $\mathcal{F}$ with an euclidean ground cost can be expressed as follow using the marginal constraints on $\mT \in \mathcal{U}(\vh,\vh)$,
	\begin{equation}\label{eq:FGW_matrix}
	\begin{split}
	\mathcal{F}(\mA_1,\mA_2,\mT) &= \sum_{ij} \|\va_{1,i} - \va_{2,j}\|_2^2 T_{ij} \\
	&= \sum_i \|\va_{1,i}\|_2^2 h_i +\sum_j \|\va_{1,j}\|_2^2 h_j - 2\sum_{ij}\scalar{\va_{1,i}}{\va_{2,j}}T_{ij}\\
	&= \scalar{\mD_{\vh}^{1/2}\mA_1}{\mD_{\vh}^{1/2}\mA_1}_F+\scalar{\mD_{\vh}^{1/2}\mA_2}{\mD_{\vh}^{1/2}\mA_2}_F -2\scalar{\mA_1\mA_2^{\top}}{\mT}_F
	\end{split}
	\end{equation}
	Returning to our main problem \ref{eq:FGW_bound1}, a straigth-forward development of its Wasserstein term $\mathcal{F}$ using \eqref{eq:FGW_matrix} leads to the following equality,
	\begin{equation}
	\begin{split}
	&\mathcal{F}\left(\widetilde{\mA}(\vw^{(1)}),\widetilde{\mA}(\vw^{(2)}), \mT\right)= \sum_{pq} \left( w_p^{(1)}w_q^{(1)} +w_p^{(2)}w_q^{(2)}\right)\scalar{\mD_{\vh}^{1/2}\overline{\mA_p}}{\mD_{\vh}^{1/2}\overline{\mA_q}}_F -2 w^{(1)}_p w^{(2)}_q\scalar{\mA_p\mA_q^{\top}}{\mT}_F
	\end{split}
	\end{equation}
	Similarly than for the proof of Proposition 1, $\mT \in \mathcal{U}(\vh,\vh)$ is an optimal admissible coupling minimizing the FGW problem, thus \eqref{eq:FGW_bound1} is upper bounded by its evaluation in the sub-optimal admissible coupling $\mD_{\vh} \in \mathbf{U}(\vh,\vh)$. Let $\mM_1=\mM=(\scalar{\mD_{\vh} \overline{\mC_p}}{\overline{\mC_q} \mD_{\vh}}_{F})_{pq}$ the PSD matrix coming from the proof of Proposition \ref{prop:embed_graph}.
	
	Let $\mM_2 = \left(\scalar{\mD_{\vh}^{1/2}A_p}{\mD_{\vh}^{1/2}A_q}_F\right)_{pq}$  which is also a PSD matrix as it can be factorized as $\mB^{\top}\mB$ with $\mB = \left(vec(\mD_{\vh}^{1/2}\mA_s)\right)_{s\in[S]} \in \R^{Nd \times S}$. 
	
	Let us denote $\forall \alpha \in (0,1)$, $\mM_{\alpha}=\alpha \mM_1 + (1-\alpha)\mM_2$ which is PSD  as convex combination of PSD matrices, hence engender a Mahalanobis distance in the embedding space. To summarize, \eqref{eq:FGW_bound2} holds $\forall \alpha \in (0,1)$,
	\begin{equation}\label{eq:FGW_bound2}
	\begin{split}
	FGW_{2,\alpha}^2\left(\widetilde{\mC}(\vw^{(1)}),\widetilde{\mA}(\vw^{(1)}),\widetilde{\mC}(\vw^{(2)}),\widetilde{\mA}(\vw^{(2)}),\vh\right)&\leq  {\vw^{(1)}}^{\top}\mM_{\alpha}\vw^{(1)} +{\vw^{(2)}}^{\top} \mM_{\alpha} \vw^{(2)} - 2{\vw^{(1)}}^\top \mM_{\alpha} \vw^{(2)}\\
	&=\| \vw^{(1)}-\vw^{(2)} \|_{\mM_{\alpha}}  \\
	\end{split}
	\square
	\end{equation}

	\subsubsection{Proposition 3. Gradients of GW \textit{w.r.t.} the weights}
	In this section we will prove the following result:
	\begin{proposition}
		\label{grad_prop}
		{Let $(\mC^1,\vh^1)$ and $(\mC^2,\vh^2)$ be two graphs. Let $\mT^{*}$ be an optimal coupling of the GW problem between $(\mC^1,\vh^1),(\mC^2,\vh^2)$. We define the following cost matrix $\mM(\mT^{*}) := \left(\sum_{kl}(C^1_{ik}- C^2_{jl})^2T^{*}_{kl} \right)_{ij}$. Let $\alphab^{*}(\mT^{*}),\betab^{*}(\mT^{*})$ be the dual variables of the following linear OT problem:}
		\begin{equation*}
		\min_{\mT \in \mathcal{U}(\vh^1,\vh^2)}\langle \mM(\mT^{*}) ,\mT\rangle_{F}
		\end{equation*}
		Then $\alphab^{*}(\mT^{*})$ (\textit{resp} $\betab^{*}(\mT^{*})$) is a subgradient of the function $GW_2^2(\mC^1,\mC^2,\ \bigcdot\ ,\vh^2)$ (\textit{resp} $GW_2^2(\mC^1,\mC^2,\vh^1, \ \bigcdot \ )$).
	\end{proposition}
	In the following $\mT\geq 0$ should be understood as $\forall i,j \ T_{ij}\geq 0$. Let $(\mC^1,\vh^1)$ and $(\mC^2,\vh^2)$ be two graphs of order $n$ and $m$ with $\mC^{1} \in S_n(\R),\mC^{2}\in S_m(\R)$ and $(\vh^{1},\vh^{2}) \in \Sigma_{n} \times \Sigma_{m}$. Let $\mT^{*}$ be an optimal solution of the GW problem \emph{i.e.} $\gw_2(\C^1,\C^2,\vh^{1},\vh^{2})=\scalar{\L(\C^1,\C^2)\otimes \mT^{*}}{\mT^{*}}_{F}$. We define $\mM(\mT^{*}):=\L(\C^1,\C^2)\otimes \mT^{*}$. We consider the problem:
	\begin{equation}
	\label{eq:linear_of_gw}
	\min_{\mT \in \mathcal{U}(\vh^1,\vh^2)} \scalar{\mM(\mT^{*})}{\mT}_{F}= \min_{\mT \in \mathcal{U}(\vh^1,\vh^2)} \scalar{\L(\C^1,\C^2)\otimes \mT^{*}}{\mT}_{F}
	\end{equation}
	We will first show that the optimal coupling for the Gromov-Wasserstein problem is also an optimal coupling for the problem \eqref{eq:linear_of_gw}, \ie\ $\min_{\mT \in \mathcal{U}(\vh^1,\vh^2)} \scalar{\mM(\mT^{*})}{\mT}_{F}=\scalar{\mM(\mT^{*})}{\mT^{*}}_{F}$. This result is based on the following theorem which relates a solution of a Quadratic Program (QP) with a solution of a Linear Program (LP):
	\begin{theorem}[Theorem 1.12 in \cite{murty-linear-1988}]
		\label{murty_theo}
		Consider the following (QP):
		\begin{equation}
		\label{eq:qp_general}
		\begin{array}{cl}{\min _{\xbf} f(\xbf)} & {=\mathbf{c} \xbf+\xbf^{T} \mathbf{Q} \xbf} \\ {\text {s.t.}} & {\mathbf{A} \xbf = \mathbf{b}},\;  {\xbf \geq 0}\end{array}
		\end{equation}
		Then if $\xbf_{*}$ is an optimal solution of \eqref{eq:qp_general} it is an optimal solution of the following (LP):
		\begin{equation}
		\label{eq:lp_general}
		\begin{array}{cl}{\min _{\xbf} f(\xbf)} & {=(\mathbf{c} + \xbf_{*}^{T}\mathbf{Q} )  \xbf} \\ {\text {s.t.}} & {\mathbf{A} \xbf = \mathbf{b}},\;  {\xbf \geq 0}\end{array}
		\end{equation}
	\end{theorem}
	
	Applying Theorem \ref{murty_theo} to our case gives exactly that: 
	\begin{equation}
	\label{eq:gpislp}
	\mT^{*} \in \argmin_{\mT \in \mathcal{U}(\vh^1,\vh^2)} \scalar{\mM(\mT^{*})}{\mT}_{F}
	\end{equation}
	since $\mT^{*}$ is an optimal solution of the GW problem and so $\min_{\mT \in \mathcal{U}(\vh^1,\vh^2)} \scalar{\mM(\mT^{*})}{\mT}_{F}=\scalar{\mM(\mT^{*})}{\mT^{*}}_{F}$.

	Now let $\alphab^{*}(\mT^{*}),\betab^{*}(\mT^{*})$ be an optimal solution to the dual problem of \eqref{eq:linear_of_gw}. Then by strong duality it implies that:
	\begin{equation}
	\min_{\mT \in \mathcal{U}(\vh^1,\vh^2)}\scalar{\mM(\mT^{*})}{\mT}_{F}=\scalar{\alphab^{*}(\mT^{*})}{\vh^1}+\scalar{\betab^{*}(\mT^{*})}{\vh^2}= \scalar{\mM(\mT^{*})}{\mT^{*}}_{F}
	\end{equation}
	Since $\scalar{\mM(\mT^{*})}{\mT^{*}}_{F}=\gw_2(\C^1,\C^2,\vh^1,\vh^2)$ we have:
	\begin{equation}\label{eq:strong_dual}
	\gw_2(\C^1,\C^2,\vh^1,\vh^2)=\scalar{\alphab^{*}(\mT^{*})}{\vh^1}+\scalar{\betab^{*}(\mT^{*})}{\vh^2}
	\end{equation}
	To prove Proposition \ref{grad_prop} the objective is to show that $\betab^{*}(\mT^{*})$ is a subgradient of $F:\qbf \rightarrow \gw(\C^1,\C^2,\vh^1,\qbf)$ (by symmetry the result will be true for $\alphab^{*}(\mT^{*})$). In other words we want to prove that:
	\begin{equation}
	\forall \qbf \in \Sigma_m, \scalar{\betab^{*}(\mT^{*})}{\qbf}-\scalar{\betab^{*}(\mT^{*})}{\vh^2}\leq F(\qbf)-F(\vh^2)
	\end{equation}
	This condition can be rewritten based on the following simple lemma:
	\begin{lemma}
		\label{lemmaone}
		The dual variable $\betab^{*}(\mT^{*})$ is a subgradient of $F: \qbf \rightarrow \gw_2(\C^1,\C^2,\vh^1,\qbf)$ if and only if:
		\begin{equation}
		\forall \qbf \in \Sigma_m, \scalar{\betab^{*}(\mT^{*})}{\qbf}{}+\scalar{\alphab^{*}(\mT^{*})}{\vh^1}{}\leq F(\qbf)
		\end{equation}
	\end{lemma}
	
	\paragraph{Proof.}
	It is a subgradient if and only if:
	\begin{equation}
	\forall \qbf \in \Sigma_m, \scalar{\betab^{*}(\mT^{*})}{\qbf}{}-\scalar{\betab^{*}(\mT^{*})}{\vh^2}{}\leq F(\qbf)-F(\vh^2)
	\end{equation}
	However using \eqref{eq:strong_dual} and the definition of $F$ we have:
	\begin{equation}
	\label{eq:strong_dual_F}
	F(\vh^2)=\scalar{\alphab^{*}(\mT^{*})}{\vh^1}{}+\scalar{\betab^{*}(\mT^{*})}{\vh^2}{}
	\end{equation}
	So overall:
	\begin{equation}
	\begin{split}
	&\scalar{\betab^{*}(\mT^{*})}{\qbf}{}-\scalar{\betab^{*}(\mT^{*})}{\vh^2}{} \leq F(\qbf)-(\scalar{\alphab^{*}(\mT^{*})}{\vh^1}{}+\scalar{\betab^{*}(\mT^{*})}{\vh^2}{}) \\ 
	&\iff \scalar{\betab^{*}(\mT^{*})}{\qbf}{}+\scalar{\alphab^{*}(\mT^{*})}{\vh^1}{}\leq F(\qbf)
	\end{split}
	\end{equation}
	$\square$
	
	In order to prove Proposition \ref{grad_prop} we have to prove that the condition in Lemma \ref{lemmaone} is satisfied. We will do so by leveraging the weak-duality of the GW problem as described in the next lemma:
	\begin{lemma}
		For any vectors $\alphab \in \R^{n},\betab \in \R^{m}$ we define: 
		\begin{equation*}
		\mathcal{G}(\alphab,\betab):=\min_{\mT \geq 0} \scalar{\L(\C^1,\C^2)\otimes \mT-\alphab \one_m^{\top}- \one_n\betab^{\top}}{\mT}{}
		\end{equation*}
		Let $\mT^{*}$ be an optimal solution of the GW problem. Consider:
		\begin{equation}
		\label{eq:eqlemmmm}
		\min_{\mT \in \mathcal{U}(\vh^1,\vh^2)} \scalar{\mM(\mT^{*})}{\mT}_{F}
		\end{equation}
		where $\mM(\mT^{*}):=\L(\C^1,\C^2)\otimes \mT^{*}$. Let $\alphab^{*}(\mT^{*}),\betab^{*}(\mT^{*})$ be the dual variables of the problem in \eqref{eq:eqlemmmm}. If $\mathcal{G}(\alphab^{*}(\mT^{*}),\betab^{*}(\mT^{*}))= 0$ then $\betab^{*}(\mT^{*})$ is a subgradient of $F: \qbf \rightarrow \gw_2(\C^1,\C^1,\vh^1,\qbf)$
	\end{lemma}
	\paragraph{Proof.}
	Let $\qbf \in \Sigma_m$ be any weights vector be fixed. Recall that $F: \qbf \rightarrow \gw_2(\C^1,\C^2,\vh^1,\qbf)$ so that:
	\begin{equation}
	\label{eq:beforelag}
	F(\qbf)=\gw_2(\C^1,\C^2,\vh^1,\qbf)=\min_{\mT \in \mathcal{U}(\vh^1,\qbf)} \scalar{\L(\C^1,\C^2)\otimes \mT}{\mT}{}
	\end{equation}
	The Lagrangian associated to \eqref{eq:beforelag} reads:
	\begin{equation}
	\label{eq:lag_cond}
	\begin{split}
	&\min_{\mT \geq 0} \max_{\alphab,\betab} \texttt{L}(\mT,\alphab,\betab) \text{ where }\texttt{L}(\mT,\alphab,\betab):= \scalar{\L(\C^1,\C^2)\otimes \mT}{\mT}{}+ \scalar{\vh^1-\mT \one_m}{\alphab}{}+ \scalar{\qbf-\mT^{\top} \one_n}{\betab}{}
	\end{split}
	\end{equation}
	Moreover by weak Lagrangian duality:
	\begin{equation}
	\label{eq:weak_duality}
	\min_{\mT \geq 0} \max_{\alphab,\betab} \texttt{L}(\mT,\alphab,\betab) \geq \max_{\alphab,\betab} \min_{\mT \geq 0}  \texttt{L}(\mT,\alphab,\betab)
	\end{equation}
	However: 
	\begin{equation*}
	\begin{split}
	\max_{\alphab,\betab} \min_{\mT \geq 0}  \texttt{L}(\mT,\alphab,\betab)&=\max_{\alphab,\betab} \scalar{\alphab}{\vh^1}{}+\scalar{\betab}{\qbf}{}+ \min_{\mT \geq 0} \scalar{\L(\C^1,\C^2)\otimes \mT-\alphab \one_m^{\top}- \one_n\betab^{\top}}{\mT}{} \\
	&=\max_{\alphab,\betab} \scalar{\alphab}{\vh^1}{}+\scalar{\betab}{\qbf}{}+\mathcal{G}(\alphab,\betab)
	\end{split}
	\end{equation*}
	So by considering the dual variable $\alphab^{*}(\mT^{*}),\betab^{*}(\mT^{*})$ defined previously we have:
	\begin{equation}
	\label{eq:duddu}
	\max_{\alphab,\betab} \min_{\mT \geq 0}  \texttt{L}(\mT,\alphab,\betab)\geq \scalar{\alphab^{*}(\mT^{*})}{\vh^1}{}+\scalar{\betab^{*}(\mT^{*})}{\qbf}{}+\mathcal{G}(\alphab^{*}(\GG^{*}),\betab^{*}(\mT^{*}))
	\end{equation}
	Now combining \eqref{eq:weak_duality} and \eqref{eq:duddu} we have:
	\begin{equation}
	\begin{split}
	\min_{\mT \geq 0} \max_{\alphab,\betab} \texttt{L}(\mT,\alphab,\betab) \geq \scalar{\alphab^{*}(\mT^{*})}{\vh^1}{}+\scalar{\betab^{*}(\mT^{*})}{\qbf}{}+\mathcal{G}(\alphab^{*}(\mT^{*}),\betab^{*}(\mT^{*}))
	\end{split}
	\end{equation}
	Since $F(\qbf)=\min_{\mT \geq 0} \max_{\alphab,\betab} \texttt{L}(\mT,\alphab,\betab)$ we have proven that:
	\begin{equation}
	\label{eq:eqeqeqeqe}
	\forall \qbf \in \Sigma_m, \scalar{\betab^{*}(\mT^{*})}{\qbf}{}+\scalar{\alphab^{*}(\mT^{*})}{\vh^1}{}+\mathcal{G}(\alphab^{*}(\mT^{*}),\betab^{*}(\mT^{*}))\leq F(\qbf)
	\end{equation} 
	However Lemma \ref{lemmaone} states that $\betab^{*}(\mT^{*})$ is a subgradient of $F$ if and only if:
	\begin{equation}
	\forall \qbf \in \Sigma_m, \scalar{\betab^{*}(\mT^{*})}{\qbf}{}+\scalar{\alphab^{*}(\mT^{*})}{\vh^1}{}\leq F(\qbf)
	\end{equation} 
	So combining \eqref{eq:eqeqeqeqe} with Lemma \ref{lemmaone} proves:
	\begin{equation}
	\mathcal{G}(\alphab^{*}(\mT^{*}),\betab^{*}(\mT^{*})) \geq 0 \implies \betab^{*}(\mT^{*}) \text{ is a subgradient of } F
	\end{equation}
	However we have $F(\vh^2)=\scalar{\alphab^{*}(\mT^{*})}{\vh^1}{}+\scalar{\betab^{*}(\mT^{*})}{\vh^2}{}$ by \eqref{eq:strong_dual_F}. So $\mathcal{G}(\alphab^{*}(\mT^{*}),\betab^{*}(\mT^{*}))\leq 0$ using \eqref{eq:eqeqeqeqe} with $\qbf=\vh^2$. So we can only hope to have $\mathcal{G}(\alphab^{*}(\mT^{*}),\betab^{*}(\mT^{*}))= 0$. 
	$\square$
	
	The previous lemma states that it is sufficient to look at the quantity $\mathcal{G}(\alphab^{*}(\mT^{*}),\betab^{*}(\mT^{*}))$ in order to prove that $\betab^{*}(\mT^{*})$ is a subgradient of $F$. Interestingly the condition $\mathcal{G}(\alphab^{*}(\mT^{*}),\betab^{*}(\mT^{*}))=0$ is satisfied which proves Proposition \ref{grad_prop} as sated in the next lemma:
	\begin{lemma}
		With previous notations we have $\mathcal{G}(\alphab^{*}(\mT^{*}),\betab^{*}(\mT^{*}))=0$. In particular $\betab^{*}(\mT^{*})$ is a subgradient of $F$ so that Proposition \ref{grad_prop} is valid. 
	\end{lemma}
	
	\paragraph{Proof.}
	We want to find:
	\begin{equation*}
	\begin{split}
	&\mathcal{G}(\alphab^{*}(\mT^{*}),\betab^{*}(\mT^{*}))= \min_{\mT \geq 0} \scalar{\L(\C^1,\C^2)\otimes \mT-\alphab^{*}(\mT^{*}) \one_m^{\top}- \one_n\betab^{*}(\mT^{*})^{\top}}{\mT}{} \\
	\end{split}
	\end{equation*}
	We define $H(\mT):=\scalar{\L(\C_1,\C_2)\otimes \mT-\alphab^{*}(\mT^{*}) \one_m^{\top}- \one_n\betab^{*}(\mT^{*})^{\top}}{\mT}{}$. Since $\mT^{*}$ is optimal coupling for $\min_{\mT \in \mathcal{U}(\vh^1,\vh^2)} \scalar{\mM(\mT^{*})}{\mT}_{F}$ by \eqref{eq:gpislp} then for all $i,j$ we have $T_{ij}^{*}(\mM(\mT^{*})_{ij}-\alpha_i^{*}(\mT^{*})-\beta_j^{*}(\mT^{*}))=0$ by the property of the optimal couplings for the Wasserstein problems. Equivalently: 
	\begin{equation}
	\begin{split}
	\forall (i,j) \in [n] \times [m], \ T_{ij}^{*}([\L(\C^1,\C^2)\otimes \mT^{*}]_{ij}-\alpha_i^{*}(\mT^{*})-\beta_j^{*}(\mT^{*}))=0
	\end{split}
	\end{equation}
	Then:
	\begin{equation}
	\begin{split}
	&H(\mT^{*})=\tr\left({\mT^{*}}^{\top}(\L(\C^1,\C^2)\otimes \mT^{*}-\alphab^{*}(\mT^{*}) \one_m^{\top}- \one_n\betab^{*}(\mT^{*})^{\top})\right) \\
	&=\sum_{ij} T^{*}_{ij} (\L(\C^1,\C^2)\otimes \mT^{*}-\alphab^{*}(\mT^{*}) \one_m^{\top}- \one_n\betab^{*}(\mT^{*})^{\top})_{ij} \\
	&=\sum_{ij} T^{*}_{ij} ([\L(\C^1,\C^2)\otimes \mT^{*}]_{ij}-\alpha_i^{*}(\mT^{*})-\beta_j^{*}(\mT^{*})) = 0
	\end{split}
	\end{equation}
	Which proves $\mathcal{G}(\alphab^{*}(\mT^{*}),\betab^{*}(\mT^{*}))=0$.
	$\square$

	\subsection{Algorithmic details}
	\subsubsection{GDL for graphs without attributes} 
	We propose to model a graph as a weighted sum of pairwise relation
	matrices. More precisely, given a graph $G=(\mC,\vh)$ and a \emph{dictionary}
	$\{\overline{\mC}_{s}\}_{s \in [S]} \subset S_N(\R)$  we want to find a linear representation $\sum_{s
		\in [S]} w_{s} \overline{\mC}_{s}$ of the graph $G$, as faithful as
	possible. The dictionary is made of pairwise relation matrices of graphs with
	order $N$. $\vw=(w_{s})_{s \in [S]} \in \Sigma_S$ is referred as
	\emph{embedding} and denotes the coordinate of the graph $G$ in the dictionary. We rely
	on the GW distance to assess the quality of our linear
	approximation and propose to minimize it to estimate its optimal embedding.

	\subsubsection{Gromov-Wasserstein unmixing}
	We first study the unmixing problem that consists in projecting a graph on the
	linear representation discussed above, \emph{i.e.} estimate the optimal
	embedding $\vw$ of a graph $G$. Our GW unmixing problem reads as 
	\begin{equation}   
	\min_{\vw \in \Sigma_S}\quad GW^2_2\left(\mC, \widetilde{\mC}(\vw)\right) - \lambda \|\vw\|^2_2 \label{eq:unmix}
	\end{equation}
	\begin{equation}
	\text{where,} \qquad \widetilde{\mC}(\vw) = \sum_s w_s \overline{\mC_s}
	\end{equation}
	where $\lambda \in \mathbb{R}^{+}$ induces a \textbf{negative} quadratic regularization promoting sparsity on the
	simplex as discussed in \citet{li2016methods}. In order to solve the non-convex problem in \eqref{eq:unmix}, we propose to
	use a Block Coordinate Descent (BCD) algorithms \citep{tseng2001convergence}. We fully detail the algorithm in the following and refer our readers to the main paper for the discussion on this approach.
	\begin{algorithm}[H]
		\caption{BCD for GW unmixing problem \ref{eq:unmix}}
		\label{alg:BCD1}
		\begin{algorithmic}[1]
			\STATE Initialize $\vw=\frac{1}{S}\mathbf{1}_S$
			\REPEAT
			\STATE Compute OT matrix $\mT$ of $GW_2^2\left(\mC, \widetilde{\mC}(\vw) \right)$, with CG algorithm ~\citep[Alg.1 \& 2]{vayer-fused-2018}.
			\STATE Compute the optimal $\vw$ solving \eqref{eq:unmix} for a fixed
			$\mT$ with CG algorithm \ref{alg:CGw}
			\UNTIL{convergence}
		\end{algorithmic}
	\end{algorithm}
	
	\begin{algorithm}[H]
		\caption{CG for solving GW unmixing problem \emph{w.r.t} $\vw$ given $\mT$}
		\label{alg:CGw}
		\begin{algorithmic}[1]
			\REPEAT
			\STATE Compute $\vg$, gradients  \emph{w.r.t} $\vw$ of $\mathcal{E}(\mC, \widetilde{\mC}(\vw),\mT)$ following \eqref{eq:gradw_GW}.
			\STATE Find direction $\vx^{\star}= \argmin_{\vx\in \Sigma_S} \vx^T\vg$ 
			\STATE Line-search: denoting $\vz(\gamma) = \gamma \vx^{\star} + (1 - \gamma)\vw$,
			\begin{equation}\label{eq:linesearchGW}
			\gamma^\star = \argmin_{\gamma \in (0,1)} \mathcal{E}(\mC,\widetilde{\mC}(\vz(\gamma)), \mT)= \argmin_{\gamma \in (0,1)} a \gamma^2 + b \gamma +c
			\end{equation}
			\STATE $\vw \leftarrow \vz(\gamma^\star)$
			\UNTIL{convergence}
		\end{algorithmic}
	\end{algorithm}
	
	Partial derivates of the GW objective $\mathcal{E}$ \emph{w.r.t} $\vw = (\frac{\partial \mathcal{E} }{\partial w_s})_{s \in [S]}$ are expressed in \eqref{eq:gradw_GW}, and further completed with gradient of the negative regularization term .
	\begin{equation} \label{eq:gradw_GW}
	\frac{\partial \mathcal{E} }{\partial w_s}(\mC,\widetilde{\mC}(\vw),\mT) = 2tr\{ \left(\overline{\mC_s} \odot \widetilde{\mC}(\vw)\right) \vh \vh^\top - \overline{\mC_s} \mT^{\top} \mC^\top \mT \}
	\end{equation}
	The coefficient of the second-order polynom involved in \eqref{eq:linesearchGW} used to solve the problem, are expressed as follow,
	\begin{equation}
	a= tr\{ \left(\widetilde{\mC}(\vx^\star - \vw) \odot \widetilde{\mC}(\vx^\star - \vw)\right)\vh\vh^T \}- \lambda \| \vx^\star- \vw\|_2^2
	\end{equation}
	\begin{equation}
	b= 2tr\{\left( \widetilde{\mC}(\vx^\star - \vw) \odot \widetilde{\mC}(\vw)\right)\vh \vh^\top - \widetilde{\mC}(\vx^\star - \vw)\mT^{\top}\mC^T \mT\}- 2 \lambda \scalar{\vw}{\vx-\vw}
	\end{equation}
	
	\subsubsection{Dictionary Learning and online algorithm}
	
	Assume now that the dictionary $\{\overline{\mC}_s\}_{s \in [S]}$ is not known
	and has to be estimated from the data.
	We define a dataset of $K$ graphs $\left\{ G^{(k)} :
	(\mC^{(k)},\vh^{(k)}) \right\}_{k \in [K]}$. Recall that each graph $G^{(k)}$ of
	order $N^{(k)}$ is summarized by its pairwise relation matrix $\mC^{(k)} \in
	S_{N^{(k)}}(\R)$ and weights $\vh^{(k)} \in \Sigma_{N^{(k)}}$ over nodes.
	The DL problem, that aims at estimating the optimal dictionary
	for a given dataset can be expressed as:
	\begin{equation}
	\min_{\begin{smallmatrix}\{\vw^{(k)}\}_{k\in [K]} \\
		\{\overline{\mC}_s\}_{s \in [S]} \end{smallmatrix}} \sum_{k=1}^K GW^2_2\left(\mC^{(k)},\widetilde{\mC}(\vw^{(k)})\right)- \lambda \|\vw^{(k)}\|^2_2 \label{eq:dl}
	\end{equation}
	where $\vw^{(k)} \in \Sigma_S, \overline{\mC}_s \in S_N(\R)$. We refer the reader to the main paper for the discussion on the non-convex problem \ref{eq:dl}. To tackle this problem we proposed a stochastic algorithm \ref{alg:GW1}
	
	\begin{algorithm}[h]
		\caption{GDL: stochastic update of atoms $\{\overline{\mC}_s\}_{s\in [S]}$}
		\label{alg:GW1}
		\begin{algorithmic}[1]
			\STATE Sample a minibatch of graphs $\mathcal{B} :=\{\mC^{(k)}\}_{k \in \mathcal{B}}$ .
			\STATE Compute optimal $\{(\vw^{(k)},\mT^{(k)})\}_{k \in [B]}$ by solving B independent unmixing problems with Alg.\ref{alg:BCD1}. 
			\STATE Projected gradient step with estimated gradients $\widetilde{\nabla}_{\overline{\mC}_s}$ (see \eqref{eq:gradCs_GW}), $\forall s \in [S]$: \vspace{-2mm}
			\begin{equation}
			\overline{\mC}_s \leftarrow Proj_{S_N(\R)}( \overline{\mC}_s - \eta_C \widetilde{\nabla}_{\overline{\mC}_s} )
			\end{equation}
		\end{algorithmic}
	\end{algorithm}
	Estimated gradients \emph{w.r.t} $\{\overline{\mC_s}\}$ over a minibatch of graphs $\mathcal{B} :=\{\mC^{(k)}\}_{k \in \mathcal{B}}$ given unmixing solutions $\{(\vw^{(k)},\mT^{(k)})\}_{k \in [B]}$ read:
	\begin{equation}\label{eq:gradCs_GW}
	\widetilde{\nabla}_{\overline{\mC_{s}}}\left(\sum_{k\in \mathcal{B}}\mathcal{E}(\mC^{(k)}, \widetilde{\mC}(\vw^{(k)}), \mT^{(k)}\right) = \frac{2}{B} \sum_{k \in \mathcal{B}} w^{(k)}_s \{ \widetilde{\mC}(\vw^{(k)}) \odot \vh \vh^\top -  \mT^{{(k)}\top}\mC^{(k)\top}  \mT^{(k)}\}
	\end{equation}
	\subsection{GDL for graph with nodes attribute}
	We can also define the same DL procedure for labeled graphs using the FGW distance. The unmixing part defined in \eqref{eq:unmix} can be adapted by considering a linear embedding of the similarity matrix \emph{and} of the feature matrix parametrized by the \emph{same} $\vw$. 
	\subsubsection{Fused Gromov-Wasserstein unmixing}
	More precisely, given a labeled graph $G=(\mC,\mA,\vh)$ (see Section \ref{sec:defs} ) and a \emph{dictionary}
	$\{(\overline{\mC_s},\overline{\mA_s})\}_{s \in [S]} \subset S_N(\R) \times \R^{N \times d}$  we want to find a linear representation $(\sum_{s
		\in [S]} w_{s} \overline{\mC_s},\sum_{s
		\in [S]} w_{s} \overline{\mA_s})$ of the labeled graph $G$, as faithful as
	possible in the sense of the FGW distance. The FGW unmixing problem that consists in projecting a labeled graph on the linear representation discussed above reads as follow, $\forall \alpha \in (0,1)$, 
	
	\begin{equation}   
	\min_{\vw \in \Sigma_S}\quad FGW^2_{2,\alpha}\left(\mC,\mA, \widetilde{\mC}(\vw),\widetilde{\mA}(\vw)\right) - \lambda \|\vw\|^2_2 \label{eq:unmix_fgw}
	\end{equation}
	\begin{equation}
	\text{where,} \quad\widetilde{\mC}(\vw) = \sum_s w_s \overline{\mC_s} \quad \text{and} \quad \widetilde{\mA}(\vw) = \sum_s w_s \overline{\mA_s}
	\end{equation}
	where $\lambda \in \mathbb{R}^{+}$. A similar discussion than for the GW unmixing problem \ref{eq:unmix} holds. We adapt the BCD algorithm detailed in \ref{alg:BCD1} to labeled graphs in Alg.\ref{alg:BCD_fgw}, to solve the non-convex problem of \eqref{eq:unmix_fgw}.
	
	\begin{algorithm}[H]
		\caption{BCD for FGW unmixing problem \ref{eq:unmix_fgw}}
		\label{alg:BCD_fgw}
		\begin{algorithmic}[1]
			\STATE Initialize $\vw=\frac{1}{S}\mathbf{1}_S$
			\REPEAT
			\STATE Compute OT matrix $\mT$ of $FGW_{2,\alpha}^2\left(\mC,\mA, \widetilde{\mC}(\vw), \widetilde{\mA}(\vw)\right)$, with CG algorithm ~\citep[Alg.1 \& 2]{vayer-fused-2018}.
			\STATE Compute the optimal $\vw$ solving \eqref{eq:unmix_fgw} for a fixed
			$\mT$ with CG algorithm \ref{alg:CGw_fgw}.
			\UNTIL{convergence}
		\end{algorithmic}
	\end{algorithm}
	\vspace{-4mm}
	\begin{algorithm}[H]
		\caption{CG for solving FGW unmixing problem \emph{w.r.t} $\vw$ given $\mT$}
		\label{alg:CGw_fgw}
		\begin{algorithmic}[1]
			\REPEAT
			\STATE Compute $\vg$, gradients  \emph{w.r.t} $\vw$ of \eqref{eq:unmix_fgw} given $\mT$ following \eqref{eq:gradw_FGW}.
			\STATE Find direction $\vx^{\star}= \argmin_{\vx\in \Sigma_S} \vx^T\vg$ 
			\STATE Line-search: denoting $\vz(\gamma) = \gamma \vx^{\star} + (1 - \gamma)\vw$,
			\begin{equation}\label{eq:linesearchGW}
			\gamma^\star = \argmin_{\gamma \in (0,1)} \alpha \mathcal{E}(\mC,\widetilde{\mC}(\vz(\gamma)), \mT) + (1-\alpha)\mathcal{F}(\mA,\widetilde{\mA}(\vz(\gamma)), \mT)= \argmin_{\gamma \in (0,1)} a \gamma^2 + b \gamma +c
			\end{equation}
			\STATE $\vw \leftarrow \vz(\gamma^\star)$
			\UNTIL{convergence}
		\end{algorithmic}
	\end{algorithm}
	
	Partial derivates of the FGW objective $\mathcal{G}_{\alpha} :=\alpha\mathcal{E} +(1-\alpha)\mathcal{F}$ \emph{w.r.t} $\vw$ are expressed in equations \ref{eq:gradw_GW} and \ref{eq:gradw_FGW}, and further completed with gradient of the negative regularization term.
	\begin{equation} \label{eq:gradw_FGW}
	\begin{split}
	\frac{\partial \mathcal{G}_\alpha }{\partial w_s}(\mC,\mA,\widetilde{\mC}(\vw),\widetilde{\mA}(\vw),\mT) &= \alpha\frac{\partial \mathcal{E} }{\partial w_s}(\mC,\widetilde{\mC}(\vw),\mT) +(1-\alpha)\frac{\partial \mathcal{F} }{\partial w_s}(\mA,\widetilde{\mA}(\vw),\mT) \\
	&= \alpha\frac{\partial \mathcal{E} }{\partial w_s}(\mC,\widetilde{\mC}(\vw),\mT) 
	+2(1-\alpha) tr\{\mD_{\vh} \widetilde{\mA}(\vw) \overline{\mA_s}^\top - \mT^\top \mA \overline{\mA_s}^\top\}
	\end{split} 
	\end{equation}
	where $\mD_{\vh} = diag(\vh)$.
	The coefficients of the second-order polynom involved in \eqref{eq:linesearchGW} used to solve the problem, satisfy the following equations,
	\begin{equation}
	a= \alpha tr\{ \left(\widetilde{\mC}(\vx^\star - \vw) \odot \widetilde{\mC}(\vx^\star - \vw)\right)\vh\vh^T \} +(1-\alpha)tr\{\mD_{\vh}\widetilde{\mA}(\vx^\star - \vw)\widetilde{\mA}(\vx - \vw)^\top\}- \lambda \| \vx^\star- \vw\|_2^2
	\end{equation}
	\begin{equation}
	\begin{split}
	b&= 2\alpha tr\{\left( \widetilde{\mC}(\vx^\star - \vw) \odot \widetilde{\mC}(\vw)\right)\vh\vh^\top - \widetilde{\mC}(\vx^\star - \vw)\mT^{\top}\mC^T \mT\}\\
	&+(1-\alpha) tr\{ \mD_{\vh}\widetilde{\mA}(\vx^\star - \vw)\widetilde{\mA}(\vw)^\top - \mT^\top \mA \widetilde{\mA}(\vx^\star - \vw)^\top\} - 2 \lambda \scalar{\vw}{\vx-\vw}
	\end{split}
	\end{equation}
	\subsubsection{Dictionary Learning and online algorithm}
	
	Assume now that the dictionary $\{(\overline{\mC}_s,\overline{\mA}_s)\}_{s \in [S]}$ is not known
	and has to be estimated from the data.
	We define a dataset of $K$ labeled graphs $\left\{ G^{(k)} :
	(\mC^{(k)},\mA^{(k)},\vh^{(k)}) \right\}_{k \in [K]}$. Recall that each labeled graph $G^{(k)}$ of
	order $N^{(k)}$ is summarized by its pairwise relation matrix $\mC^{(k)} \in
	S_{N^{(k)}}(\R)$, its matrix of node features $\mA^{(k)} \in \R^{N^{(k)}\times d}$ and weights $\vh^{(k)} \in \Sigma_{N^{(k)}}$ over nodes.
	The DL problem, that aims at estimating the optimal dictionary
	for a given dataset can be expressed as:
	\begin{equation}
	\min_{\begin{smallmatrix}\{\vw^{(k)}\}_{k\in [K]} \\
		\{(\overline{\mC_s},\overline{\mA_s})\}_{s \in [S]} \end{smallmatrix}} \sum_{k=1}^K FGW^2_{2,\alpha}\left(\mC^{(k)},\mA^{(k)},\widetilde{\mC}(\vw^{(k)}),\widetilde{\mA}(\vw^{(k)})\right)- \lambda \|\vw^{(k)}\|^2_2 \label{eq:dl_fgw}
	\end{equation}
	where $\vw^{(k)} \in \Sigma_S, \overline{\mC}_s \in S_N(\R), \overline{\mA}_s \in \R^{N \times d}$. We refer the reader to the main paper for the discussion on the non-convex problem \ref{eq:dl} which can be transposed to problem \ref{eq:dl_fgw}. To tackle this problem we proposed a stochastic algorithm \ref{alg:FGW1}
	
	\begin{algorithm}[h]
		\caption{GDL: stochastic update of atoms $\{(\overline{\mC_s},\overline{\mA_s})\}_{s\in [S]}$}
		\label{alg:FGW1}
		\begin{algorithmic}[1]
			\STATE Sample a minibatch of graphs $\mathcal{B} :=\{(\mC^{(k)}, \mA^{(k)})\}_{k \in \mathcal{B}}$ .
			\STATE Compute optimal $\{(\vw^{(k)},\mT^{(k)})\}_{k \in [B]}$ by solving B independent unmixing problems with Alg.\ref{alg:BCD_fgw}. 
			\STATE Gradients step with estimated gradients $\widetilde{\nabla}_{\overline{\mC_s}}$ (see \eqref{eq:gradCs_GW}), and $\widetilde{\nabla}_{\overline{\mA_s}}$ (see \eqref{eq:gradAs_fgw}), $\forall s \in [S]$. : \vspace{-2mm}
			\begin{equation}
			\overline{\mC_s} \leftarrow Proj_{S_N(\R)}( \overline{\mC_s} - \eta_C \widetilde{\nabla}_{\overline{\mC_s}} )
			\qquad \text{and} \qquad
			\overline{\mA_s} \leftarrow  \overline{\mA_s} - \eta_A \widetilde{\nabla}_{\overline{\mA_s}} 
			\end{equation}
			
		\end{algorithmic}
	\end{algorithm}
	Estimated gradients \emph{w.r.t} $\{\overline{\mC_s}\}$ and $\{\overline{\mA_s}\}$ over a minibatch of graphs $\mathcal{B} :=\{(\mC^{(k)},\mA^{(k)})\}_{k \in \mathcal{B}}$ given unmixing solutions $\{(\vw^{(k)},\mT^{(k)})\}_{k \in [B]}$ can be computed separately. The ones related to the GW objective are described in \eqref{eq:gradCs_GW}, while the ones related to the Wasserstein objective satisfy \eqref{eq:gradAs_fgw}:
	\begin{equation}\label{eq:gradAs_fgw}
	\widetilde{\nabla}_{\overline{\mA_{s}}}\left(\sum_{k\in \mathcal{B}}\mathcal{F}(\mA^{(k)}, \widetilde{\mA}(\vw^{(k)}), \mT^{(k)})\right) = \frac{2}{B} \sum_{k \in \mathcal{B}}w^{(k)}_s \{\mD_{\vh}\widetilde{\mA}(\vw^{(k)}) - \mT^\top \mA^{(k)}\}
	\end{equation}
	\subsection{Learning the graph structure and nodes distribution}
	Here we extend our GDL model defined in
	equation \ref{eq:dl} and propose to learn atoms of the form $\{\overline{\mC_s},\overline{\vh_s}\}_{s\in [S]}$. In this setting we have
	two independent dictionaries modeling the relative importance of the nodes with
	$\overline{\vh_s} \in \Sigma_N$, and their pairwise relations through $\overline{\mC_s}$. This
	dictionary learning problem reads:
	\begin{equation}
	\min_{\substack{\{(\vw^{(k)},\vv^{(k)})\}_{k \in [K]}\\
			\{(\overline{\mC}_s,\overline{\vh}_s)\}_{s\in [S]}}} \sum_{k=1}^K  GW^2_2\left(\mC^{(k)},\widetilde{\mC}(\vw^{(k)}), \vh^{(k)}, \widetilde{\vh}(\vv^{(k)})\right) - \lambda \|\vw^{(k)}\|^2_2 - \mu \|\vv^{(k)}\|^2_2\label{eq:dl_h}
	\end{equation}
	where $\vw^{(k)},\vv^{(k)} \in \Sigma_S$ are the structure and distribution embeddings and the linear models are defined as:
	\begin{equation}\label{eq:GDLextended_rpz}
	\forall k,\
	\widetilde{\vh}(\vv^{(k)}) = \sum_s v^{(k)}_s\overline{\vh_s},\quad  
	\widetilde{\mC}(\vw^{(k)}) = \sum_s w^{(k)}_s \overline{\mC_s} 
	\end{equation}
	Here we exploit fully the GW
	formalism by estimating simultaneously the graph distribution $\widetilde{\vh}$
	and its geometric structure $\widetilde{\mC}$. Optimization problem \ref{eq:dl_h} can
	be solved by an adaptation of stochastic Algorithm \ref{alg:GW1}. Indeed, in the light of the proposition \ref{grad_prop}, we can derive the following \eqref{eq:unmix_h1} between the input graph $(\mC^{(k)}, \vh^{(k)})$ and its embedded representation $\widetilde{\mC}(\vw^{(k)}$ and $\widetilde{\vh} (\vv^{(k)})$, given an optimal coupling $\mT^{(k)}$ satisfying Proposition \ref{grad_prop}, 
	\begin{equation}\label{eq:unmix_h1}
	2\scalar{\mL(\mC^{(k)},\widetilde{\mC}(\vw^{(k)})) \otimes \mT^{(k)}}{\mT^{(k)}}\\ 
	= \scalar{\vu^{(k)}}{\vh^{(k)}} +  \scalar{\widetilde{\vu}^{(k)}}{\widetilde{\vh}(\vv^{(k)})}
	\end{equation}
	where $\vu^{(k)}, \widetilde{\vu}^{(k)}$ are dual potentials of the induced linear OT problem.
	
	First, with this observation we estimate
	the structure/node weights unmixings
	$(\vw^{(k)},\vv^{(k)})$ for the graph $G^{(k)}$. We proposed the BCD algorithm \ref{alg:BCD_extended} derived from the initial BCD \ref{alg:BCD1}. Note that the dual variables of the induced linear OT problems are centered to ensure numerical stability. 
	
	\begin{algorithm}[H]
		\caption{BCD for extended GW unmixing problem inherent to \eqref{eq:dl_h}}
		\label{alg:BCD_extended}
		\begin{algorithmic}[1]
			\STATE Initialize embeddings such as $\vw=\vv= \frac{1}{S}\mathbf{1}_S$
			\REPEAT
			\STATE Compute OT matrix $\mT$ of $GW_2^2\left(\mC, \widetilde{\mC}(\vw),\vh,\widetilde{\vh}(\vv) \right)$, with CG algorithm ~\citep[Alg.1 \& 2]{vayer-fused-2018}.  From the finale iteration of CG, get dual potentials $(\vu, \widetilde{\vu})$ of the corresponding linear OT problem (see Proposition \ref{grad_prop}).
			\STATE Compute the optimal $\vv$ by minimizing \eqref{eq:unmix_h1} \emph{w.r.t} $\vv$ given $\widetilde{\vu}$ with a CG algorithm. 
			\STATE Compute the optimal $\vw$ solving \eqref{eq:unmix} given 
			$\mT$ and $\vv$ with CG algorithm \ref{alg:CGw}. 
			\UNTIL{convergence}
		\end{algorithmic}
	\end{algorithm}
	
	Second, now that we benefit from an algorithm to project any graph $G^{(k)} = (\mC^{(k)},\vh^{(k)})$ onto the linear representations described in \ref{eq:GDLextended_rpz}, we extend the stochastic algorithm \ref{alg:GW1}. to the problem \ref{eq:dl_h}. This extension is described in algorithm \ref{alg:GW_extended}. 
	
	\begin{algorithm}[h]
		\caption{extended GDL: stochastic update of atoms $\{(\overline{\mC_s},\overline{\vh_s})\}_{s\in [S]}$}
		\label{alg:GW_extended}
		\begin{algorithmic}[1]
			\STATE Sample a minibatch of graphs $\mathcal{B} :=\{(\mC^{(k)},\vh^{(k)})\}_{k \in \mathcal{B}}$ .
			\STATE Compute optimal embeddings $\{(\vw^{(k)},\vv^{(k)})\}_{k \in [B]}$ coming jointly with the set of OT variables $(\mT^{(k)},\vu^{(k)}, \widetilde{\vu}^{(k)})$ by solving B independent unmixing problems with Alg.\ref{alg:BCD_extended}. 
			\STATE Projected gradient step with estimated gradients $\widetilde{\nabla}_{\overline{\mC_s}}$ (see \eqref{eq:gradCs_GW}) and $\widetilde{\nabla}_{\overline{\vh_s}}$ (see \eqref{eq:gradhs_GW}), $\forall s \in [S]$: \vspace{-2mm}
			\begin{equation}
			\overline{\mC_s} \leftarrow Proj_{S_N(\R)}( \overline{\mC_s} - \eta_C \widetilde{\nabla}_{\overline{\mC_s}} ) \qquad \text{and} \qquad \overline{\vh_s} \leftarrow Proj_{\Sigma_N}( \overline{\vh_s} - \eta_h \widetilde{\nabla}_{\overline{\vh_s}} )
			\end{equation}
		\end{algorithmic}
	\end{algorithm}
	For a minibatch a graphs $\{\mC_k,\vh_k\}_{k \in [B]}$, once each unmixing problems are solved independently estimating unmixings $\{(\vw^{(k)}, \vw^{(k)})\}_k$ and the underlying OT matrix $\mT^{(k)}$ associated with potential $\widetilde{\vu}^{(k)}$, we perform simultaneously a projected gradient step
	update of $\{\overline{\mC}_s\}_s$ and $\{\overline{\vh}_s\}_s$. The estimated gradients of \eqref{eq:dl_h} \emph{w.r.t} $\{\overline{\vh_s} \}_s$ reads $\forall s \in [S]$,
	\begin{equation}\label{eq:gradhs_GW}
	\widetilde{\nabla}_{\overline{\vh_s}} \cdot = \frac{1}{2B} \sum_{k\in [B]}v^{(k)}_s \widetilde{u}^{(k)}
	\end{equation}
	
	\subsection{Numerical experiments}
	
	\subsubsection{Datasets}
	\vspace{-4mm}
	\begin{table}[!h]
		\centering
		\caption{Datasets descriptions}
		\label{tab:data}
		\scalebox{0.7}{
			\begin{tabular}{l|r|r|r|r|r|r|r|r}
				\hline
				datasets &  features &  \#graphs & \#classes & mean \#nodes  &  min \#nodes & max \#nodes & median \#nodes & mean connectivity rate\\ \hline
				IMDB-B &     None &  1000   &  2 & 19.77 & 12 & 136 & 17 & 55.53\\ \hline
				IMDB-M &     None & 1500 & 3 & 13.00 &  7 & 89 & 10 & 86.44 \\ \hline
				MUTAG & $\{0..2\} $ & 188 & 2 & 17.93&10 & 28 & 17.5 & 14.79\\ \hline
				PTC-MR& $\{0,..,17\}$ & 344 & 2 &  14.29& 2 & 64 & 13 & 25.1 \\ \hline
				BZR& $\R^3$ & 405 & 2 & 35.75& 13 & 57 & 35 & 6.70\\ \hline
				COX2& $\R^3$ &467& 2 &41.23& 32& 56 & 41 & 5.24 \\ \hline
				PROTEIN& $\R^{29}$ & 1113& 2 & 29.06& 4 & 620& 26 & 23.58\\ \hline
				ENZYMES& $\R^{18}$ & 600 & 6& 32.63 & 2 & 126 & 32 & 17.14\\ \hline
		\end{tabular}}
	\end{table}
	We considered well-known benchmark datasets divided into three categories: i) IMDB-B and IMDB-M~\citep{yanardag-deep-2015} gather graphs without node attributes derived from social networks; ii) graphs with discrete attributes representing chemical compounds from MUTAG~\citep{debnath1991structure} and cuneiform signs from PTC-MR~\citep{krichene2015efficient}; iii) graphs with real vectors as attributes, namely  BZR, COX2~\citep{sutherland2003spline}
	and PROTEINS, ENZYMES~\citep{borgwardt2005shortest}. Details on each dataset are reported in Table \ref{tab:data}
	\subsubsection{Settings}
	In the following, we detail the benchmark of our methods on supervised classification along additional (shared) considerations we made regarding the learning of our models.
	To consistently benchmark methods and configurations, as real graph datasets commonly used in machine learning literature show a high variance considering structure, we perform a nested cross validation (using 9 folds for training, 1 for testing, and reporting the average accuracy of this experiment repeated 10 times) by keeping same folds across methods. All splits are balanced \emph{w.r.t} labels. In following results, parameters of SVM are cross validated within $C \in \{10^{-7}, 10^{-6}, ..., 10^7\}$ and $\gamma \in \{2^{-10},2^{-9},...,2^{10}\}$.
	
	For our approach, similar dictionaries are considered for unsupervised classification presented in the main paper, than for the supervised classification benchmark detailed in the following. So we refer the reader to the main paper for most implementation details. For completeness, we picked a batch size of 16. We initialized learning rate on the structure $\{\overline{C}_s\}$ at 0.1. In the presence of node features, we set a learning rate on $\{\overline{A}_s\}$ of $0.1$ if $\alpha<0.5$ and $1.0$ otherwise. We optimized our dictionaries without features over 20 epochs and those with features over 40 epochs. In the following,  we denote GDL-w the SVMs derived from embeddings $\vw$ endowed with the Mahalanobis distance. While GDL-g denotes the SVMs derived from embedded graphs with the (F)GW distance. \citep{xu_gromov-wasserstein_2019} proposed a supervised extension to their Gromov-Wasserstein Factorization (GWF), we refer to GWF-r and GWF-f when the dictionary atoms have random size or when we fix it to match our method. His supervised approach consists in balancing the dictionary objective with a classification loss by plugging a MLP classifier to the unconstrained embedding space. We explicitly regularized the learning procedure by monitoring the accuracy on train splits. Note that in their approach they relaxed constraints of their unmixing problems by applying a softmax on unconstrained embeddings to conduct barycenters estimation. Moreover, they constrain the graph atoms to be non-negative as it enhances numerical stability of their learning procedure. For fair comparisons, we considered this restriction for all dictionaries even if we did not observe any noticeable impact of this hypothesis on our approach. As for unsupervised experiments, we followed their architecture choices. We further validated their regularization coefficient in $\{1.,0.1,0.01,0.001\}$. Their model converge over 10 epochs for datasets without features, and 20 epochs otherwise.

	We also considered several kernel based approaches. (FGWK) The kernels $e^{-\gamma FGW}$ proposed by \citep{vayer-fused-2018}  where pairwise distances are computed using CG algorithms using POT library \citep{flamary2017pot}. To get a grasp of the approximation error from this algorithmic approach, we also applied the MCMC algorithm proposed by  \citep{chowdhury-generalized-2020} to compute FGW distance matrices with a better precision (S-GWK). As the proper graph representations for OT-based methods is still a question of key interest, we consistently benchmarked our approach and these kernels when we consider adjacency and shortest-path representations. Moreover, we experimented on the heat kernels over normalized laplacian matrices suggested by \citep{chowdhury-generalized-2020} on datasets without attributes, where we validated the diffusion parameter $t \in
	\{5,10,20\}$. We also reproduced the benchmark
	for classification on Graph Kernels done by \citep{vayer-fused-2018} by keeping their tested parameters for each method. (SPK)
	denotes the shortest path kernel \citep{borgwardt2005shortest}, (RWK) the
	random walk kernel \citep{gartner2003graph}, (WLK) the Weisfeler Lehman kernel
	\citep{vishwanathan2010graph}, (GK) the graphlet count kernel
	\citep{shervashidze09}. For real valued vector attributes, we
	consider the HOPPER kernel (HOPPERK) \citep{feragen2013scalable} and the
	propagation kernel (PROPAK) \citep{neumann2016propagation} . We built upon the
	GraKel library \citep{siglidis2020grakel}  to construct the kernels. 
	
	Finally to compare our performances to recent state-of-the-art models for supervised graph classification, we partly replicated the benchmark done by \citep{xu2018powerful}. We experimented on their best model GIN-0 and the model of \citep{niepert-learning-2016} PSCN. r. For both we used the Adam optimizer \citep{kingma2014adam} with initial learning rate 0.01 and decayed the learning rate by 0.5 every 50 epochs. The number of hidden units is chosen depending on dataset statistics as they propose, batch normalization \citep{ioffe2015batch} was applied on each of them. The batch size was fixed at 128. We fixed a dropout ratio of 0.5 after the dense layer \citep{srivastava2014dropout}. The number of epochs was 150 and the model with the best cross-validation accuracy averaged over the 10 folds was selected at each epoch.
	
	\subsubsection{Results on supervised classification} The accuracies of the nested-cross validation on described datasets are reported in Tables \ref{tab:res1}, \ref{tab:res2}, \ref{tab:res3}. First, we observe as anticipated that the model GIN-0 \citep{xu2018powerful} outperforms most of the time other methods including PSCN, which has been consistently argued in their paper. Moreover, (F)GW kernels over the embedded graphs built thanks to our dictionary approach consistently outperforms (F)GW kernels from input graphs. Hence, it supports that our dictionaries are able to properly denoise and capture discriminant patterns of these graphs, outperforming other models expect GNN on 6 datasets out of 8. The Mahalanobis distance over embeddings $\vw$ demonstrates satisfying results compared to FGWK relatively to the model simplification it brings. We also observe consistent improvements of the classification performances when we use the MCMC algorithm \citep{chowdhury-generalized-2020} to estimate (F)GW pairwise distance matrices, for all tested graph representations reported. This estimation procedure for (F)GW distances is computationally heavy compared to the usual CG gradient algorithm \citep{vayer-fused-2018}. Hence, we believe that it could bring significant improvements to our dictionary learning models but would increase too consequently the run time of solving unmixing problems required for each dictionary updates. Finally, results over adjacency and shortest path representations interestingly suggest that their suitability \emph{w.r.t} (F)GW distance is correlated to the averaged connectivity rate (see \ref{tab:data}) in different ways depending on the kind of node features. We envision to study these correlations in future works.
	\vspace{-4mm}
	\begin{table}[!h]
		\caption{\textbf{Graphs without attributes:} Classification results of 10-fold nested-cross validation on real datasets. Best results are highlighted in bolt independently of the depicted model category, and the best performances from not end-to-end supervised methods are reported in italic. }
		\label{tab:res1}
		\begin{center}
			\scalebox{0.81}
			{\begin{tabular}{|l|r|r|r|}
					\hline
					category & model & IMDB-B & IMDB-M \\ \hline
					OT (Ours) & GDL-w (ADJ) &70.11(3.13)& 49.01(3.66) \\ 
					& GDL-g (ADJ) & \textit{72.06(4.09)} & \textit{50.64(4.41)} \\
					& GDL-w (SP) & 65.4(3.65) & 48.03(3.80) \\
					& GDL-g (SP) & 68.24(4.38) & 48.47(4.21) \\ \hline 			
					OT & FGWK (ADJ) & 70.8(3.54) &  48.89(3.93)\\  
					& FGWK (SP) & 65.0(3.69)& 47.8(3.84)\\
					& FGWK (heatLAP) & 67.7(2.76)& 48.11(3.96) \\
					& S-GWK (ADJ) & 71.95(3.87) & 49.97(3.95)\\
					& S-GWK (heatLAP) & 71.05(3.02) & 49.24(3.49)\\
					& GWF-r (ADJ) & 65.08(2.85) & 47.53(3.16)\\
					& GWF-f (ADJ) & 64.68(2.27) & 47.19(2.96) \\ \hline
					Kernels & GK (K=3) & 57.11(3.49) & 41.85(4.52) \\
					& SPK & 56.18(2.87) & 39.07(4.89) \\ \hline
					
					GNN & PSCN & 71.23(2.13)& 45.7(2.71)\\
					& GIN-0 & \textbf{74.7(4.98)}& \textbf{52.19(2.71)}\\ \hline
			\end{tabular}}
		\end{center}
	\end{table}
	\vspace{-5mm}
	\begin{table}[!h]
		\centering
		\caption{\textbf{Graphs with discrete attributes :} Classification results of 10-fold nested-cross validation on real datasets with discrete attributes (one-hot encoded). Best results are highlighted in bolt independently of the depicted model category, and the best performances from not end-to-end methods are reported in italic.}
		\label{tab:res2}
		\scalebox{0.81}
		{\begin{tabular}{|l|r|r|r|}
				\hline
				category & model & MUTAG &PTC-MR \\ \hline
				OT (Ours) & GDL-w (ADJ) & 81.07(7.81) & 55.26(8.01)\\
				& GDL-g (ADJ) & 85.84(6.86) & 58.45(7.73)\\
				& GDL-w (SP)  & 84.58(6.70)& 55.13(6.03)\\
				& GDL-g (SP)  & \textit{87.09(6.34)}& 57.09(6.59)\\ \hline
				OT & FGWK (ADJ)  & 82.63(7.16)& 56.17(8.85) \\  
				& FGWK (SP)  & 84.42(7.29)& 55.4(6.97)\\
				& S-GWK (ADJ)  & 84.08(6.93)& 57.89(7.54)\\
				& GWF-r (ADJ)  & -& -\\
				& GWF-f (ADJ)  & -& -\\ \hline
				Kernels & GK (K=3)  & 82.86(7.93)& 57.11(7.24) \\
				& SPK &  83.29(8.01)&  60.55(6.43)\\
				& RWK & 79.53(7.85)& 55.71(6.86)\\
				& WLK & 86.44(7.95)& \textit{63.14(6.59)}\\ \hline
				GNN & PSCN & \textbf{91.4(4.41)} & 58.9(5.12)\\
				& GIN-0  & 88.95(4.91)& \textbf{64.12(6.83)}\\ \hline
		\end{tabular}}
	\end{table}
	\vspace{-5mm}
	\begin{table}[!h]
		\centering
		\caption{\textbf{Graphs with vectorial attributes:} Classification results of 10-fold nested-cross validation on real datasets with vectorial features. Best results are highlighted in bolt independently of the depicted model category, and the best performances from not end-to-end supervised methods are reported in italic.}
		\label{tab:res3}
		\scalebox{0.81}
		{\begin{tabular}{|l|r|r|r|r|r|r|}
				\hline
				category & model & BZR & COX2 & ENZYMES & PROTEIN \\  \hline
				OT (ours) & GDL-w (ADJ)& 87.32(3.58)& 76.59(3.18) & 70.68(3.36) & 72.13(3.14)\\ 
				& GDL-g (ADJ) &  \textit{87.81(4.31)}& 78.11(5.13) & 71.44(4.19) & 74.59(4.95)\\ 
				& GDL-w (SP) &  83.96(5.51)& 75.9(3.81) & 69.95(5.01) & 72.95(3.68)\\ 
				& GDL-g (SP) &  84.61(5.89)& 76.86(4.91) & 71.47(5.98)& \textit{74.86(4.38)}\\ \hline
				OT& FGWK (ADJ) & 85.61(5.17) & 77.02(4.16) & 72.17(3.95) & 72.41(4.70)\\
				& FGWK (SP) & 84.15(6.39) & 76.53(4.68) & 70.53(6.21)& 74.34(3.27)\\
				&S-GWK (ADJ)& 86.91(5.49) & 77.85(4.35) & \textbf{73.03(3.84)} & 73.51(4.96)\\
				& GWF-r (ADJ) & 83.61(4.96) & 75.33(4.18) & 72.53(5.39)& 73.64(2.48)\\
				& GWF-f (ADJ) & 83.72(5.11) & 74.96(4.0) & 72.14(4.97)& 73.06(2.06)\\
				\hline
				
				Kernels & HOPPERK& 84.51(5.22) & \textit{79.68(3.48)}& 46.2(3.75) & 72.07(3.06)\\
				& PROPAK& 80.01(5.11) & 77.81(3.84)& 71.84(5.80)& 61.73(4.5)\\ \hline
				GNN & PSCN& 83.91(5.71) & 75.21(3.29)& 43.89(3.91) & 74.96(2.71)\\
				& GIN-0& \textbf{88.71(5.48)} & \textbf{81.13(4.51)}& 68.6(3.69) & \textbf{76.31(2.94)}\\ \hline
		\end{tabular}}
	\end{table}
	\newpage
	\subsubsection{Complementary results on unsupervised classification}
	\paragraph{vanilla GDL} As mentioned in section 4 of the main paper, we considered a fixed batch size for learning our models on labeled graphs, which turned out to be a limitation for the dataset ENZYMES. We report in table \ref{tab:clustering_enzymes} our models performance on this dataset for a batch size fixed to 64 instead of 32 within the framework detailed above. These results are consistent with those observed on the other datasets.
	\begin{table*}[h!]
		\vspace*{-5mm}
		\caption{Clustering : dataset ENZYMES}
		\label{tab:clustering_enzymes}
		\begin{center}
			\scalebox{0.81}{
				\begin{tabular}{|c|c|}
					\hline
					MODELS & ENZYMES \\
					\hline
					GDL &  71.83(0.18)\\
					$\text{GDL}_\lambda $& $\mathbf{72.92(0.28)}$\\ 
					\hline
			\end{tabular}}
		\end{center}
	\end{table*}
	
	\vspace*{-5mm}
	\paragraph{extended version of GDL} We report here a companion study for clustering tasks which further supports our extension of GDL to the learning of node weights. As there is no Mahalanobis upper-bound for the linear models learned with this extension as their node weights are a priori different, we compare performances of K-means with GW distance applied on the embedded graphs produced with vanilla GDL, the extended version of GDL denoted here $\text{GDL}_{h}$ and GWF. Similar considerations have been made for learning $\text{GDL}_{\vh}$ than those detailed for GDL, and we completed these results with an ablation of the quadratic negative regularization parameterized by $\lambda$. Results provided in \ref{tab:clustering_GWkmeans} show that GW Kmeans applied to the graph representations from our method $\text{GDL}_h$ leads to state-of-the-art performances.
	
	%\textbf{Details on experiments:}\\
	\begin{table}[h!]
		\vspace*{-5mm}
		\caption{Clustering: RI from GW Kmeans on embedded graphs.}
		\label{tab:clustering_GWkmeans}
		\begin{center}
			\scalebox{0.81}{
				\begin{tabular}{|l|c|c|c|}
					\hline
					models & $\lambda$ & IMDB-B & IMDB-M\\
					\hline
					GDL \text{ } (ours)&  0 & 51.54(0.29)& 55.86(0.25) \\
					& $>0$ & 51.97(0.48) & 56.41(0.35) \\ \hline
					%			GDL+dist \emph{(embed)} & 0 &51.09(0.18) & 54.89(0.22) \\
					%			 & $>0$ &51.31(0.41) & 54.81(0.19) \\\hline
					GDL$_\mathbf{h}$  (ours)& 0& 52.51(0.22) & $\mathbf{57.12(0.3)}$ \\ 
					& $>0$& $\mathbf{53.09(0.38)}$ & 56.95(0.25) \\
					\hline 
					GWF-r & NA &51.39(0.15) & 55.80(0.21)\\
					GWF-f & NA &50.93(0.39) & 54.48(0.26)\\
					\hline
			\end{tabular}}%\vspace{-2mm}
		\end{center}
	\end{table}
	
	\subsubsection{Runtimes}
	We report in Table \ref{warp-tab:runtimes} averaged runtimes for the same relative precision of $10^{-4}$ to compute one graph embedding on learned dictionaries from real datasets.
	\begin{table}[h!]\vspace*{-5mm}
		\caption{Averaged runtimes.}
		\label{warp-tab:runtimes}
		\begin{center}
			\scalebox{0.8}{
				\begin{tabular}{|c|c|c|c|}
					\hline
					dataset & \# atoms & GDL  &  GWF \\
					\hline
					IMDB-B & 12 & 52 ms & 123 ms\\
					& 16 & 69 ms & 186 ms \\ \hline
					IMDB-M & 12 & 44 ms & 101 ms\\
					& 18 & 71 ms & 168 ms \\ \hline
			\end{tabular}}
		\end{center}
	\end{table}
\end{document}